\documentclass[runningheads]{llncs}

\usepackage{accv}

\usepackage{accvabbrv}

\usepackage{graphicx}
\usepackage{booktabs}
\usepackage{wrapfig}
\usepackage{enumitem}
\usepackage{colortbl}
\usepackage[table]{xcolor}
\usepackage{pifont}  
\usepackage{placeins}
\definecolor{style}{RGB}{172,170,170}
\definecolor{appearance}{RGB}{144,218,117}
\definecolor{scene}{RGB}{119,174,226}

\usepackage[accsupp]{axessibility}  

\usepackage{hyperref}
\usepackage{fontawesome5}
\usepackage{orcidlink}

\begin{document}

\title{Not Just a Subject: Capturing Adaptive Scene-Specific Variance for Visual Storytelling} 

\titlerunning{Not Just a Subject}

\author{SeungJu Cha\inst{1}\orcidlink{0009-0003-1941-0691} \and
Ye-Chan Kim\inst{1}\orcidlink{0009-0003-0809-0047} \and
Kwanyoung Lee\inst{1}\orcidlink{0009-0006-4525-871X} \and
Dong-Jin Kim\inst{1}\orcidlink{0000-0001-7231-7494}}

\authorrunning{S. Cha et al.}

\institute{Hanyang University, Seoul 04763, South Korea \\
\email{\{sju9020, dpcksdl78, mobled37, djdkim\}@hanyang.ac.kr}\\
\faGithub\ \url{https://github.com/SeungJuCha/EOS-Story} }

\maketitle

\begin{abstract}
Recent advancements in diffusion models have significantly improved text-to-image generation, enabling their application to visual storytelling, where coherent image sequences are generated from narrative prompts. However, existing methods primarily focus on identity preservation and often fail to adapt to diverse scene variations, resulting in semantic misalignment and structural inflexibility. We propose EOS-Story, a training-free framework that harmonizes identity consistency with scene-specific adaptation across text and vision spaces. 
In the text space, we decouple the narrative into identity and scene sub-prompts to prevent information dilution, but this leaves the scene without access to appearance cues.
To bridge this gap, we introduce Cross-Contextual Bridging, which softly injects a phrase-level appearance residual into the scene's $\langle$EOS$\rangle$ embedding. In the vision space, we find that scale mismatch between identity and scene subjects causes a few identity tokens to be repeatedly transferred, producing structural artifacts. We address this with Adaptive Identity Preservation, which anchors identity features at the scene-specific scale (Identity Anchoring) and integrates them via Identity-Aware Self-Attention. To evaluate performance under varied spatial compositions, we extend the Consistory+ benchmark with diverse framing scenarios (Consistory+\_+). Extensive experiments show that EOS-Story achieves competitive or superior performance on both benchmarks.
\end{abstract}

\section{Introduction}
\label{sec:intro}

Diffusion models~\cite{rombach2022high,podell2023sdxl} have recently advanced the quality of text-to-image generation, enabling diverse applications in visual content creation. 
Among these, the visual storytelling task, applicable to digital comics and character-driven animation~\cite{phung2025cineverse}, is particularly challenging because it requires aligning with evolving narrative semantics while maintaining a coherent subject across scenes~\cite{tewel2024training}.

\begin{figure}[!t]
    \captionsetup{type=figure}
\includegraphics[width=0.98\linewidth]{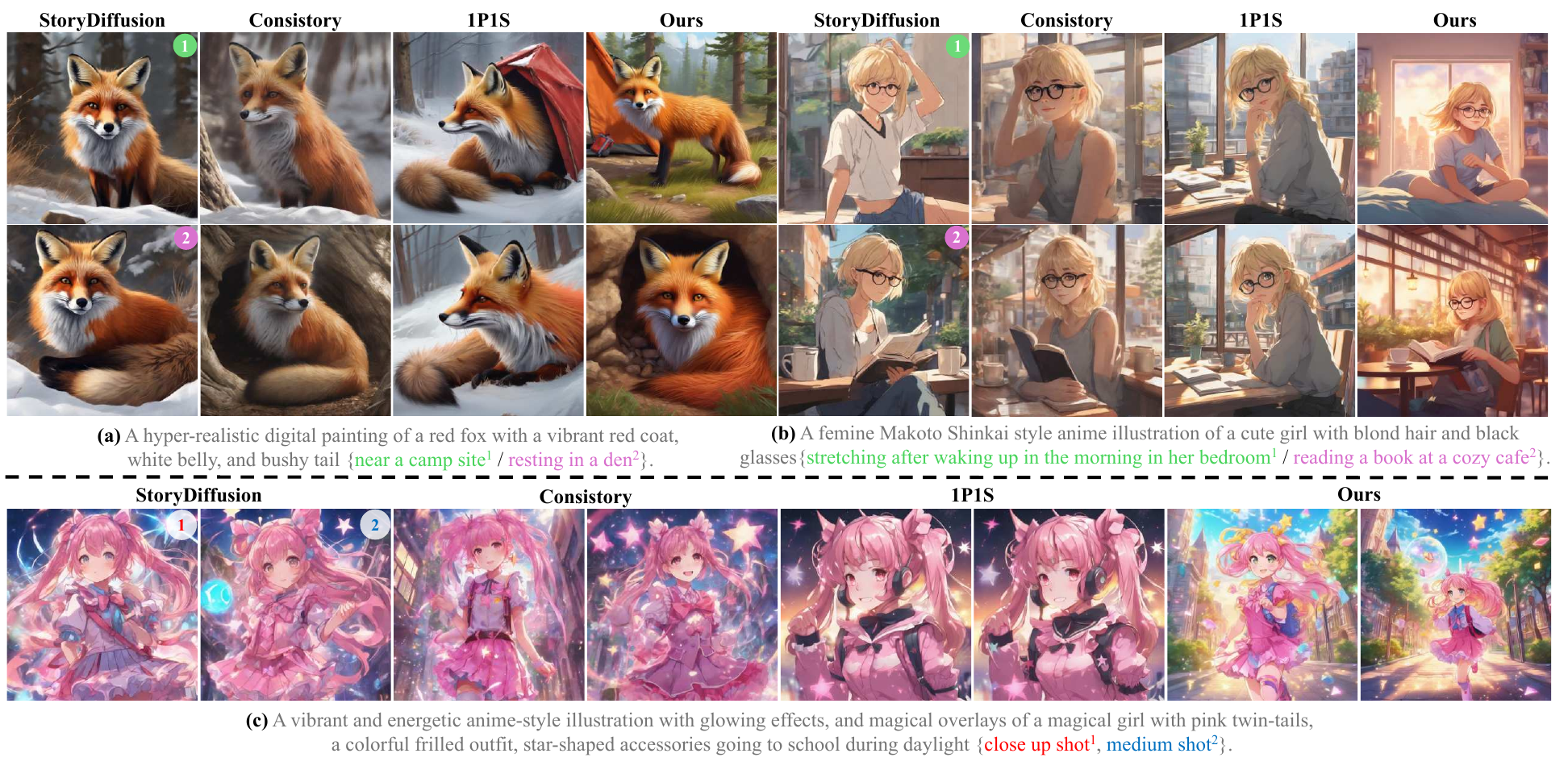}
\caption{\textbf{Limitations in scene-level adaptation of previous methods.} Existing approaches often fail to account for \textbf{scene-specific variance}: (a) \textbf{semantic misalignment} in simple cases, (b) \textbf{loss of fine-grained prompt details} in complex scenarios, and (c) inability to adapt to \textbf{diverse viewpoints and scales}. These issues often lead to repetitive subject structures and weakened alignment with scene-specific prompts. In contrast, our method achieves better scene-level fidelity by adaptively harmonizing identity consistency with the unique context of each scene.}
    \label{fig:text_misalignment}
\end{figure}
\vspace{-1mm}
Recent approaches~\cite{zhou2024storydiffusion, liu2025one, tewel2024training} 
explicitly design mechanisms to preserve subject identity by transferring subject-relevant features into each scene during generation. 
While these methods effectively maintain identity consistency, 
they often underrepresent \textit{scene-specific variance} in both text and vision spaces—the diverse semantic and visual elements that vary across narrative contexts (\cref{fig:text_misalignment}).
First, at the semantic level, current methods~\cite{liu2025one, zhou2024storydiffusion, tewel2024training} often treat narrative prompts as monolithic inputs, where various semantics (e.g., appearance and scene descriptions) are encoded into a unified embedding. 
This can lead to \textbf{information dilution}, making it difficult for the model to uniformly attend to all details~\cite{lian2023llm}. 
As a result, even when a prompt contains rich details (e.g., specific actions or environmental changes), the model often fails to selectively extract scene-specific cues from the prompt.
This directly undermines narrative fidelity, producing scenes that ignore their intended semantics (\cref{fig:text_misalignment}a,b).
Second, at the visual level, existing approaches typically rely on spatially blind feature injection, where identity features are transferred without accounting for the spatial constraints or framing of each scene. 
This oversight of structural scene variance (e.g., camera angles or subject scales) can lead to \textbf{subject inflexibility}, such as contextually mismatched poses or repetitive structural patterns across scenes (\cref{fig:text_misalignment}b,c). 
Consequently, subjects are rendered with rigid structural patterns that ignore each scene's intended layout, further weakening alignment with scene-specific prompts.
Therefore, achieving coherent storytelling requires scene-specific adaptation, where identity consistency is harmonized with the semantic and structural variance of each scene.

To address this issue, we propose \textbf{EOS-Story}, a framework designed to adaptively account for scene-specific variance via two complementary modules,
\textbf{(1) Cross-Contextual Bridging:}  
We hypothesize that semantic misalignment arises when long and complex prompts are encoded holistically, causing scene-varying attributes to be diluted~\cite{liu2024improving,lian2023llm}.
Rather than encoding all information in a single prompt, we decouple the narrative into an identity prompt describing the subject and a scene prompt describing the scenario.
For instance, ``a woman with blond hair reading a book.'' is split into an identity prompt (``a woman with blond hair.'') and a scene prompt (``a woman reading a book.'').
However, this separation introduces a new challenge: since the scene prompt contains no appearance information, the subject generated within the scene has no way to reflect the actual target identity. 
To bridge this semantic gap without reverting to monolithic encoding, we preserve the scene sub-prompt as a semantic anchor and softly inject appearance information at the embedding level.
Motivated by the strong generative influence and summarization ability of $\langle$EOS$\rangle$ embeddings~\cite{mun2024addressing, yi2024towards}, 
we interpret it not merely as a terminal token, but as a \textit{residual-based semantic control} point for scene-aware storytelling (\cref{sec:eos_emp}).
Unlike prior EOS-based methods that substitute the $\langle$EOS$\rangle$ embedding between prompts differing by a single word (e.g., an object class)~\cite{hu2024token,mun2024addressing}, we capture a phrase-level appearance difference
and infuse it into the scene's $\langle$EOS$\rangle$, reflecting rich multi-attribute cues without explicit word-level repetition (\cref{sec:eosinfusion}).
This provides the scene image with the specified appearance cues at the semantic level, enabling more effective feature-level identity transfer.

\textbf{(2) Adaptive Identity Preservation:}  
To mitigate subject inflexibility induced by spatially blind feature transfer, we introduce a spatially adaptive identity injection mechanism that better aligns identity features with the scene-specific layout (\cref{sec:identityanchoring}).
We observe that the scale mismatch between identity and scene subjects prevents various identity tokens from corresponding to the scene tokens. 
As a result, only a few identity tokens are repeatedly transferred to the scene and cause structural artifacts.
Motivated by this, we apply Identity Anchoring, which \textit{adaptively resizes} identity features to match the intended framing and scale of the scene.
This enforces structural compatibility between identity and scene representations, reducing scale-induced distortions (\cref{fig:image_abla}b).
We then refine the integration through Identity-Aware Self-Attention, which enables scene features to selectively attend to subject features, building upon~\cite{tewel2024training}.
Unlike existing spatially blind transfer strategies~\cite{tewel2024training, liu2025one, zhou2024storydiffusion}, our approach supports scene-conditioned variation in subject structure (e.g., viewpoint changes) while preserving core identity characteristics.
To evaluate scene-specific structural variance, we introduce Consistory+\_+, extending Consistory+~\cite{liu2025one} with diverse framing prompts~\cite{liu2025shotbench} to simulate the realistic framing shifts of real-world storytelling.
We validate our method on Consistory+ and Consistory+\_+, demonstrating consistently improved performance under framing variations.
Our contributions are as follows: 
\vspace{-2mm}
\begin{itemize}
 \item We show that the $\langle$EOS$\rangle$ embedding can carry phrase-level, 
    multi-attribute information, enabling appearance cues to be infused during generation.
    
    \item We propose \textbf{EOS-Story}, a training-free framework with two modules: \textbf{Cross-Contextual Bridging} (text-space semantic recomposition) 
    and \textbf{Adaptive Identity Preservation} (vision-space scale-aware transfer).    
    
    \item We introduce \textbf{Consistory+\_+}, an extended benchmark with diverse 
    framing phrases, and demonstrate consistent improvements over existing 
    approaches on both the original and extended benchmarks.
\end{itemize}

\section{Related Work}
\label{sec:relatedwork}

\noindent\textbf{Visual Storytelling.}
With the advancement of diffusion models~\cite{rombach2022high,podell2023sdxl}, their applications have expanded into various domains.
To meet users’ needs for generating images that preserve subject identity, numerous personalization methods have been proposed~\cite{avrahami2024chosen, ye2023ip, li2024photomaker, gal2022image, ruiz2023dreambooth, li2023blip}.
However, these approaches primarily focus on preserving subject appearance~\cite{tewel2024training}, often overlooking how the subject should adapt across different visual contexts.
To bridge this gap, visual storytelling~\cite{zhou2024storydiffusion, liu2025one, rahman2023make, mao2024story, wang2024oneactor} has emerged as a compelling paradigm for generating identity-consistent images across diverse and dynamic scenes~\cite{liu2025one}.
Some methods fine-tune diffusion models to explicitly capture subject identity~\cite{wang2024oneactor,rahman2023make}, while~\cite{zhou2024storydiffusion, tewel2024training, liu2025one, mao2024story, Wang_2025_ICCV, gao2025subject} explore training-free strategies that leverage intermediate features from the attention mechanism.
For example, Consistory~\cite{tewel2024training} directly maps subject identity across different scenes to achieve identity consistency, while 1P1S~\cite{liu2025one} leverages the context consistency of CLIP~\cite{radford2021learning} and transfers this information through the cross-attention mechanism.
However, these methods prioritize identity consistency at the cost of scene adaptation, often producing rigid subject depictions and weak prompt alignment.
While concurrent works\cite{gao2025subject, Wang_2025_ICCV} attempt to alleviate subject rigidity, they still delegate prompt alignment to the underlying diffusion model.
In contrast, we find that faithful scene adaptation depends not only on the vision space but also on the text space, and design our framework to jointly address both—improving prompt alignment through semantic-level residual injection and adapting to structural variance through scale-aware feature transfer.


\noindent\textbf{$\langle$EOS$\rangle$ Embedding in Diffusion Models.}
In CLIP~\cite{radford2021learning}, which is widely used to encode text in text-to-image generation models, the $\langle$EOS$\rangle$ token encapsulates a contextual summary of the entire prompt~\cite{yi2024towards}. Recent mechanistic analyses~\cite{toker2025padding} further show that padding tokens, including $\langle$EOS$\rangle$, carry non-trivial semantic influence in text-to-image generation.
Built upon this, \cite{yi2024towards} demonstrates that $\langle$EOS$\rangle$ has dominant influence in image generation: replacing the $\langle$EOS$\rangle$ embedding of an original prompt (e.g., ``A blue bird'') with that of a target prompt (e.g., ``A blue chair'') can steer the generated image toward the target semantics.
Motivated by this property, several works~\cite{hu2024token, mun2024addressing, li2024get} have utilized $\langle$EOS$\rangle$ or text-embedding manipulation for downstream tasks.
Specifically, \cite{hu2024token} substitutes $\langle$EOS$\rangle$ to resolve semantic binding issues in object–attribute prompts, \cite{mun2024addressing} injects the entire base prompt embedding along with $\langle$EOS$\rangle$ into the position of specific objects to perform image editing, and \cite{li2024get} manipulates text embeddings to suppress unwanted content.
Unlike prior methods that either substitute $\langle$EOS$\rangle$ between prompts differing by a single word~\cite{hu2024token,mun2024addressing} or manipulate embeddings for content removal~\cite{li2024get}, we use $\langle$EOS$\rangle$ as a scene-aware interface to transfer phrase-level appearance information into scene-specific prompt representations for visual storytelling.
\section{Method}
\label{sec:method}

\begin{figure*}[!t]
    \centering
    \includegraphics[width=0.97\textwidth]{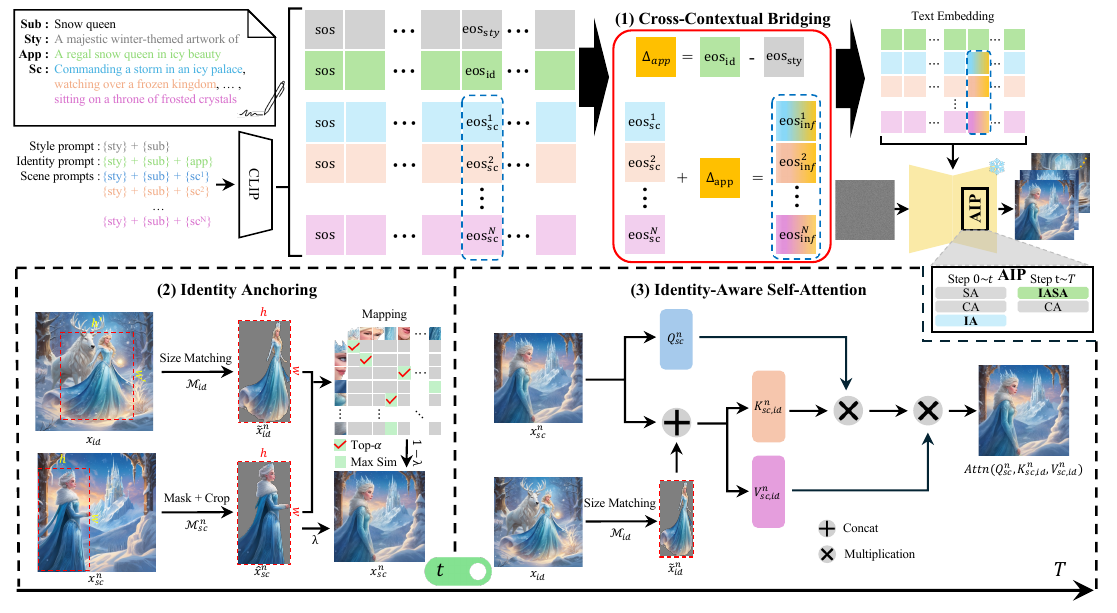}
    \caption{\textbf{EOS-Story pipeline.}
Our framework contains two complementary modules:
(1) \textbf{Cross-Contextual Bridging} for representation-level semantic recomposition (\cref{sec:eosinfusion}), and 
(2) \textbf{Adaptive Identity Preservation (AIP)} for spatially-aware structural alignment (\cref{sec:identityanchoring}).
AIP consists of two subcomponents: \textbf{Identity Anchoring} (IA), which establishes scale- and context-aware correspondence, and \textbf{Identity-Aware Self-Attention} (IASA), which refines context-dependent feature integration.
During the denoising process over diffusion timesteps $T$, we switch from explicit structural anchoring (IA) to attention-based refinement (IASA) at timestep $t$ within the unet.}
    \label{fig:pipe}
\end{figure*}

\noindent\textbf{Overview.} 
To achieve identity consistency while allowing subjects to adapt to diverse scenes, we propose \textbf{EOS-Story} (\cref{fig:pipe}), a framework that models scene-specific variance in two complementary spaces.
(1) \textbf{Cross-Contextual Bridging (CCB)} addresses semantic variance by transferring phrase-level appearance cues into the scene context through the $\langle$EOS$\rangle$ embedding, mitigating information dilution in long and complex prompts.
(2) \textbf{Adaptive Identity Preservation (AIP)} addresses structural variance by aligning identity features to each scene's subject scale (Identity Anchoring), followed by Identity-Aware Self-Attention refinement.

\subsection{EOS Embedding as a Semantic Handle}
\label{sec:eos_emp}
In this section, we explore the generative functional role of the $\langle$EOS$\rangle$ embedding.
\begin{wrapfigure}[27]{r}{0.5\textwidth}
\vspace{-15pt}
\centering
\includegraphics[width=0.48\textwidth]{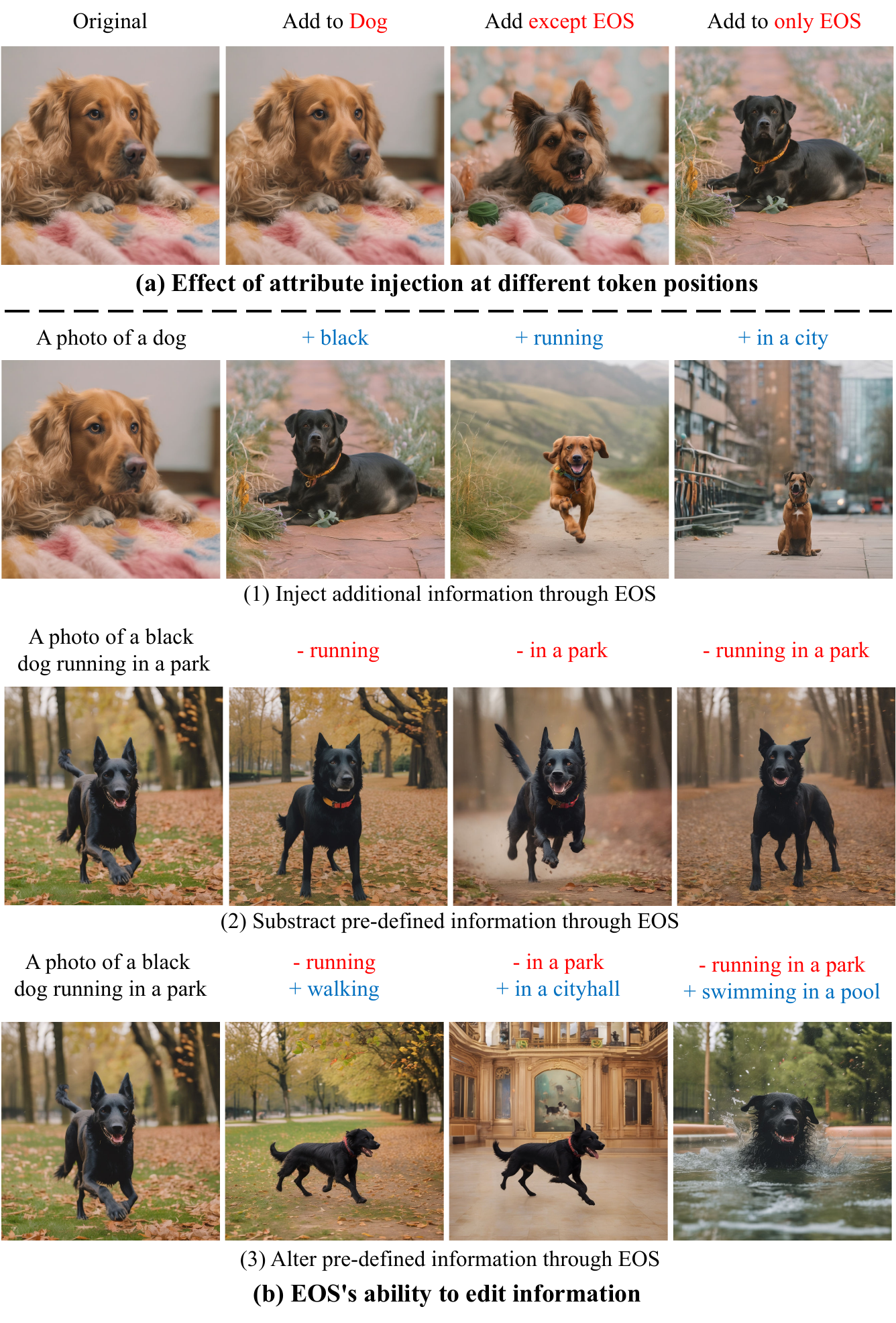}
\caption{
\textbf{Effect of EOS modification.} 
(a) Injecting ``black'' into the $\langle$EOS$\rangle$ token most effectively reflects the intended attribute.  
(b) Editing $\langle$EOS$\rangle$ embedding induces coherent semantic changes in images.
}
\label{fig:eos_token}
\end{wrapfigure}
We first test whether the $\langle$EOS$\rangle$ embedding can serve as a global semantic control point.
As shown in \cref{fig:eos_token}a, adding an attribute embedding (e.g., ``black'') at different token positions of a base prompt (``A photo of a dog'') produces notably different outcomes.
Directly adding the attribute embedding to the target word token (``dog'') fails to alter the output, and even a broad injection across all non-EOS tokens produces only minor changes.
In contrast, injecting the attribute only into the $\langle$EOS$\rangle$ embedding induces a clear semantic shift, suggesting that the $\langle$EOS$\rangle$ embedding serves as an effective handle for embedding-level semantic editing during generation.

We further examine whether semantic changes can be composed at the $\langle$EOS$\rangle$ level through additive and subtractive operations (\cref{fig:eos_token}b).
Simple subtraction often yields incomplete edits, likely because unchanged word-level tokens continue to preserve the original semantics (\cref{fig:eos_token}b2), whereas addition and replacement more reliably induce coherent semantic shifts.
Notably, our observation is that the $\langle$EOS$\rangle$ embedding can capture not only a single attribute~\cite{hu2024token,mun2024addressing} but also \textbf{multiple semantic factors}, simultaneously affecting actions and environments through one EOS-level edit (\cref{fig:eos_token}b3).
These edits occur without modifying the word-level tokens, indicating that semantic adjustments at the $\langle$EOS$\rangle$ level can override or complement persistent token-level semantics.
These observations suggest that the $\langle$EOS$\rangle$ embedding provides a context-preserving handle that can encode rich, multi-attribute semantics through a residual operation. 
Motivated by this, we design Cross-Contextual Bridging, which transfers phrase-level appearance information into each scene's $\langle$EOS$\rangle$ embedding, allowing subject-specific appearance cues to be integrated while maintaining the scene context.

\subsection{Cross-Contextual Bridging}
\label{sec:eosinfusion}
To avoid treating a story prompt as a single monolithic description, we construct reference subprompts from structured storyboard inputs (e.g., Consistory+ benchmark~\cite{liu2025one}).
Each input is decomposed into four semantic components:
\textbf{\textcolor{style}{style}}, \textbf{subject}, \textbf{\textcolor{appearance}{appearance}}, and \textbf{\textcolor{scene}{scene}}.
Encoding all components as a single prompt (e.g., ``\textcolor{style}{A majestic winter-themed artwork of} a \textcolor{appearance}{regal} snow queen \textcolor{appearance}{in icy beauty} \textcolor{scene}{watching over a frozen kingdom.}'') can dilute fine-grained appearance cues in long and compositionally complex descriptions~\cite{lian2023llm}.
To recover such appearance cues without lengthening the scene prompt, we construct three reference subprompts around a minimal shared context:
(1) \textbf{Style Prompt}: \textcolor{style}{Style} + Subject, 
(2) \textbf{Identity Prompt}: \textcolor{style}{Style} + Subject + \textcolor{appearance}{Appearance}, and 
(3) \textbf{Scene Prompt}: \textcolor{style}{Style} + Subject + \textcolor{scene}{Scene}.
The shared \textcolor{style}{Style} + Subject prompt serves as the minimal anchor for measuring appearance-specific displacement before transferring it to each scene prompt.
This anchor is scene-independent, while still preserving the subject slot and visual style, making it suitable for defining an appearance cue that can be shared across all scenes.

By keeping the scene prompt free of explicit appearance phrases, we preserve the scene description as the primary semantic context.
Cross-Contextual Bridging then reintroduces the missing appearance cue at the embedding level.
This allows the subject appearance to be recovered without modifying the word-level scene description.
Building on our empirical observations (\cref{sec:eos_emp} and Appendix~\ref{sec:eos_linear}), we utilize the $\langle$EOS$\rangle$ embedding as a controllable semantic handle, enabling phrase-level appearance transfer without increasing the textual length of the scene description.
We obtain the corresponding $\langle$EOS$\rangle$ embeddings, $eos_{sty}$, $eos_{id}$, and $eos_{sc}$, from the CLIP~\cite{radford2021learning} text encoder.
Although $eos_{id}$ contains the desired appearance information, directly adding it to $eos_{sc}$ can redundantly reinforce the shared style and subject semantics, leading to unintended semantic drift.
We therefore define the appearance residual as the local displacement from the shared style-subject anchor to the identity prompt, following the approximate linearity of CLIP embedding space~\cite{trager2023linear}: $\Delta_{app} = eos_{id} - eos_{sty}$.
We transport this residual to the scene prompt by constructing the infused embedding $eos_{inf}$:
\begin{equation}
\label{eq:eos}
eos_{inf} = eos_{sc} + \Delta_{app}.
\end{equation}
We replace each scene $\langle$EOS$\rangle$ embedding $eos_{sc}$ with its infused counterpart $eos_{inf}$.
The identity prompt embedding and the modified scene prompt embedding are then fed into the diffusion model for the subsequent feature-level identity transfer.
This formulation restores subject appearance cues within the scene prompt representation while preserving the original scene semantics.
This appearance-aware representation further reduces the semantic gap between the scene subject and its identity counterpart before feature injection, providing a stronger basis for the subsequent feature-level identity transfer (\cref{sec:identityanchoring}).

\subsection{Adaptive Identity Preservation}
\label{sec:identityanchoring}
Unlike spatially blind identity injection, we focus on anchoring identity features according to each scene's subject structure, allowing the subject to adapt in scale and viewpoint while preserving its core identity cues.
To achieve natural subject alignment across scenes, we transfer visual features from the identity image $x_{id} \in \mathbb{R}^{HW \times C}$ to $N$ scene images $\{x_{sc}^n\}_{n=1}^N \in \mathbb{R}^{N \times HW \times C}$.
We use intermediate features from the first decoder block of the U-Net, where coarse spatial structure is known to emerge~\cite{tang2023emergent}, to establish correspondence between identity and scene subject regions.
Identity Anchoring is applied during early denoising stages, when global layout is primarily established, so that later steps can refine fine-grained details without disrupting the aligned structure.
During generation, we obtain subject masks $\mathcal{M}_{id} \in \mathbb{R}^{HW\times 1}$ and $\{\mathcal{M}_{sc}^n\}_{n=1}^N \in \mathbb{R}^{N\times HW\times 1}$ from the cross-attention maps of the corresponding subject tokens.

\begin{wrapfigure}[19]{r}{0.5\textwidth}
\vspace{-15pt}
\centering
\includegraphics[width=0.48\textwidth]{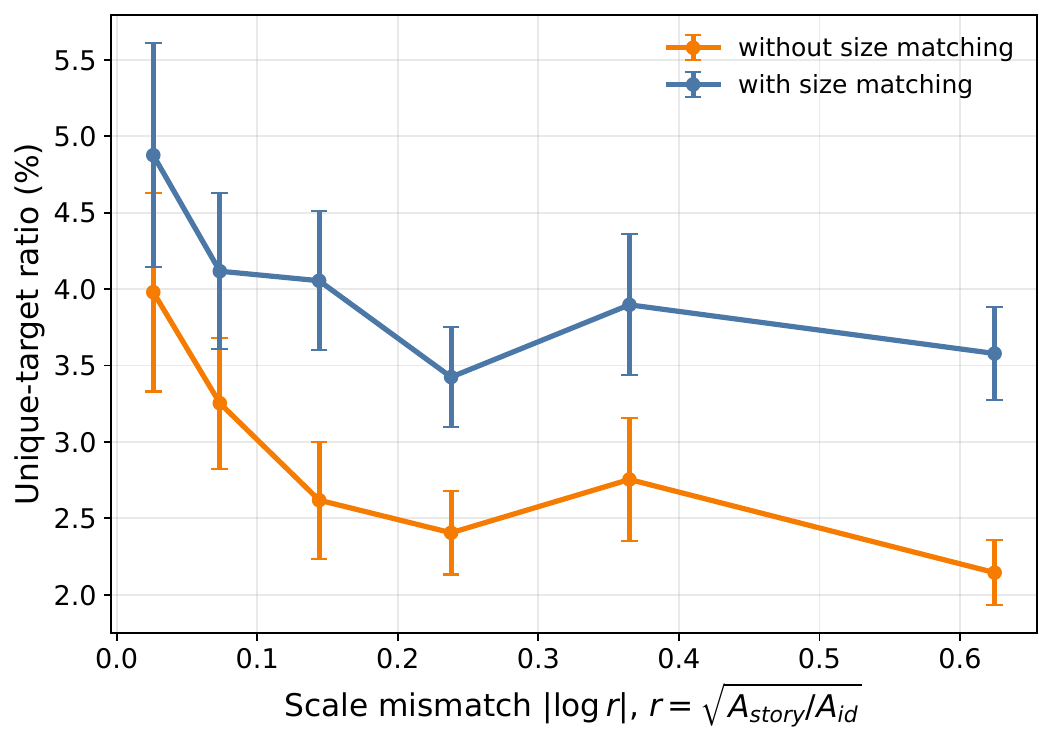}
\caption{
\textbf{Identity-token coverage.} Unique identity-token coverage—the fraction of distinct identity tokens referenced by scene tokens—drops sharply as scale mismatch $|\log r|$ grows without size matching (orange), while size matching keeps it stable (blue). Low coverage means few tokens are repeatedly transferred.}
\label{fig:mismatch}
\end{wrapfigure}

\noindent\textbf{Identity Anchoring.} 
Because each scene is guided by a different narrative prompt, the subject's scale and framing naturally vary across scenes.
We empirically find that this scale variation directly degrades feature correspondence in \cref{fig:mismatch}. 
In the figure, as the scale mismatch between identity and scene subjects grows, the correspondence from scene tokens concentrates on a smaller subset
of identity tokens.
Consequently, a few identity tokens are overused across the scene subject, causing structural artifacts and scene-inflexible depictions (\cref{fig:image_abla}b).
This suggests that establishing reliable correspondence requires first aligning the two subjects to a comparable size.

We therefore rescale the identity features to align with the subject size in each scene, enabling adaptive feature matching across varying spatial contexts (\cref{fig:pipe} (2)).
For each scene, we crop the identity subject region $\hat{x}_{id} \in \mathbb{R}^{h'w' \times C}$ and mask $\hat{\mathcal{M}}_{id} \in \mathbb{R}^{h'w' \times 1}$ using the bounding box of $\mathcal{M}_{id}$.
We similarly crop the scene subject region $\hat{x}_{sc}^n = \mathrm{Crop}(\mathcal{M}_{sc}^n \odot x_{sc}^n) \in \mathbb{R}^{hw \times C}$, where $h'w'$ and $hw$ denote the identity and scene-specific subject sizes, respectively.
We then resize the masked identity feature to match the scene subject's resolution, obtaining the size-aligned identity feature:
$\tilde{x}_{id}^n = \mathrm{Resize}^{(n)}\!\big(\hat{\mathcal{M}}_{id} \odot \hat{x}_{id}\big) \in \mathbb{R}^{hw \times C}$,
where $\mathrm{Resize}^{(n)}(\cdot)$ resizes the input to the spatial size of $\hat{x}_{sc}^n$.
This size-aware alignment allows scene features to reference identity information at a comparable scale, reducing mismatched transfers from irrelevant identity regions.
To transfer identity information into each scene, we compute a correspondence map between the size-aligned identity feature $\tilde{x}_{id}^n \in \mathbb{R}^{hw \times C}$ and the scene subject feature $\hat{x}_{sc}^n \in \mathbb{R}^{hw \times C}$.
After computing the similarity matrix $S = \hat{x}_{sc}^n \cdot (\tilde{x}_{id}^n)^\top$, we identify the most similar position in the identity feature for each scene position $i\in[1,hw]$:
\begin{equation}
\Phi_i = \mathop{\mathrm{argmax}}_{j \in [1, hw]} S_{ij}.
\end{equation}
Since every scene position is matched to its most similar identity feature, positions without a genuine counterpart are still forced into a match, propagating spurious identity cues.
To suppress unreliable transfers, we retain only matches whose similarity exceeds an adaptive confidence threshold derived from the similarity distribution.
Specifically, we set the threshold to $\mu + \sigma$, where $\mu$ and $\sigma$ denote the mean and standard deviation of similarities across all scene positions.
We then blend the scene subject and matched identity features:
\begin{equation}
\hat{x}_{sc}^n \leftarrow \lambda \cdot \hat{x}_{sc}^n + (1-\lambda)\cdot \tilde{x}_{id}^n[\Phi],
\end{equation}
where $\lambda \in [0,1]$ balances scene context preservation and identity transfer.
Finally, we replace the subject region in $x_{sc}^n$ with the blended subject feature $\hat{x}_{sc}^n$.
This adaptive alignment harmonizes identity cues with scene-specific structure, preserving the intended composition of each scene.

\noindent\textbf{Identity-Aware Self-Attention.}
Although the blended scene feature $x_{sc}^n$ incorporates visual information from the identity image $x_{id}$, explicit feature replacement may still fail to blend naturally into the surrounding scene context.
Built upon~\cite{tewel2024training}, we enable each scene to dynamically attend to identity features during self-attention (\cref{fig:pipe} (3)).  
After extracting size-aligned key and value $\tilde{K}_{id}^n$ and $\tilde{V}_{id}^n$ from the identity image feature $x_{id}$, we concatenate them with the corresponding key and value from the scene features $x_{sc}^n$: $K_{sc,id}^n = \mathrm{Concat}(K_{sc}^n, \tilde{K}_{id}^n)$, 
$V_{sc,id}^n = \mathrm{Concat}(V_{sc}^n, \tilde{V}_{id}^n)$.
The self-attention is then computed:
\begin{equation}
\mathrm{Attn}(Q^n_{sc}, K_{sc,id}^n, V_{sc,id}^n) 
= \mathrm{Softmax}\left(\frac{Q^n_{sc} K_{sc,id}^{n\top}}{\sqrt{d}}\right) V_{sc,id}^n.
\end{equation}

\section{Experiments}

\noindent\textbf{Implementation Details}. 
We use the SDXL-base-1.0 checkpoint for all experiments and set the number of denoising steps to 50.
We extract binary mask through the Otsu algorithm~\cite{otsu1975threshold}, which is commonly used~\cite{tewel2024training}.
Adaptive Identity Preservation is applied only to the decoder layers of the unet, 
with a transition from Identity Anchoring to Identity-Aware Self-Attention at step $t{=}9$, 
motivated by the observation that global layout forms in early denoising steps 
while fine-grained details emerge later~\cite{yi2024towards}.
The weighting coefficient $\lambda$ is fixed at 0.55 in all experiments.
All experiments are conducted on an NVIDIA A6000 GPU.

\noindent\textbf{Benchmarks.}
Consistory+~\cite{liu2025one} spans subjects across diverse categories (e.g., humans, animals, nature).
To evaluate generation performance under more complex spatial contexts, we introduce an extended version, Consistory+\_+, by adding diverse framing phrases covering ``shot size,'' ``camera angle,'' and ``composition.''
We first collect framing-related descriptors from prior scene-structure studies~\cite{liu2025shotbench}, filtering out those unlikely to be visually reflected in generated images.
To automatically assign appropriate descriptors, we instruct GPT-4o~\cite{hurst2024gpt} to classify each subject and apply framing tags only to animate ones (e.g., humans, animals), for which such cues are perceptually more salient and semantically aligned~\cite{Rooney2018WatchingMC}.
Further details are provided in the supplementary material.

\begin{table*}[!t]
\centering
\caption{\textbf{Quantitative results on Consistory+~\cite{liu2025one} and our extended Consistory+\_+.} Our method achieves better text alignment without compromising image quality. For StoryDiffusion~\cite{zhou2024storydiffusion}, we generate images at a resolution of $768\times768$.}
\resizebox{\textwidth}{!}{
\begin{tabular}{l|ccccc|ccccc}
\hline \hline
\multicolumn{1}{l|}{\rule{0pt}{2.6ex}} 
& \multicolumn{5}{c|}{\textbf{Consistory+}} 
& \multicolumn{5}{c}{\textbf{Consistory+\_+}} \\[0.6ex]
\cline{2-11}
\multicolumn{1}{l|}{} & CLIP-T$\uparrow$ & CLIP-I$\uparrow$ & DreamSim$\downarrow$ & Aesthetic$\uparrow$ & ImageReward$\uparrow$ & CLIP-T$\uparrow$ & CLIP-I$\uparrow$ & DreamSim$\downarrow$ & Aesthetic$\uparrow$ & ImageReward$\uparrow$ \\ \hline
SDXL (Base) & 0.8971 & 0.8430 & 0.3124 & 6.1227 & 0.8289 & 0.9071 & 0.8376 & 0.3246 & 6.1508 & 1.1835 \\ \hline

\multicolumn{11}{c}{\textcolor{red}{\scriptsize \textbf{\textit{Training-Based}}}} \\ \hline

\rowcolor{gray!15}
PhotoMaker~\cite{li2024photomaker} & 0.8871 & 0.8342 & 0.3479 & 5.6051 & 0.8289 & 0.8776 & 0.8300 & 0.3594 & 5.6222 & 0.8717 \\
\rowcolor{gray!15}
IP-Adapter~\cite{ye2023ip} & 0.8771 & 0.9398 & 0.1396 & 5.8172 & 0.8283 & 0.8351 & 0.9410 & 0.1329 & 5.8367 & 0.5114 \\ \hline

\multicolumn{11}{c}{\textcolor{green!60!black}{\scriptsize \textbf{\textit{Training-Free}}}} \\ \hline

StoryDiffusion~\cite{zhou2024storydiffusion} & \underline{0.8985} & \textbf{0.8909} & 0.2284 & 6.0884 & 0.9579 & 0.8900 & \textbf{0.8960} & \underline{0.2253} & 6.0926 & 1.1256 \\
Consistory~\cite{tewel2024training} & 0.8868 & 0.8806 & 0.2355 & \underline{6.1070} & \underline{0.9814} & \underline{0.8974} & 0.8730 & 0.2463 & \underline{6.1588} & \underline{1.1879} \\
1P1S~\cite{liu2025one} & 0.8871 & 0.8756 & \underline{0.2132} & 6.0296 & 0.8289 & 0.8900 & 0.8679 & 0.2368 & 6.1106 & 0.9731 \\
\textbf{EOS-Story (Ours)} & \textbf{0.9042} & \underline{0.8850} & \textbf{0.2005} & \textbf{6.1764} & \textbf{1.1680} & \textbf{0.8998} & \underline{0.8787} & \textbf{0.2146} & \textbf{6.1791} & \textbf{1.3217} \\
\hline \hline
\end{tabular}
}
\label{tab:main_table}
\end{table*}

\noindent\textbf{Metrics.} 
We evaluate text-to-image alignment using CLIP-T~\cite{taited2023CLIPScore}.
For subject similarity, we employ CLIP-I~\cite{radford2021learning} and DreamSim~\cite{fu2023dreamsim}, following~\cite{gao2025subject, liu2025one}.
To measure aesthetic quality, we use a SigLIP-based predictor~\cite{zhai2023sigmoid, discus2024aestheticv25}, as each story involves diverse artistic styles and appearances.
To assess fine-grained prompt details, we use ImageReward~\cite{xu2023imagereward}, a human-aligned metric capturing nuanced text-image alignment beyond superficial similarity.
To evaluate framing fidelity, we employ ShotVL-7B~\cite{liu2025shotbench} to classify whether generated images align with the framing phrases specified in the text prompt. 

\subsection{Main Experiments}


\noindent{\textbf{Quantitative Result}.} We conduct comprehensive quantitative evaluations on Consistory+ and Consistory+\_+ in \cref{tab:main_table}.
Our method consistently achieves improved performance across nearly all metrics, especially showing notable gains in CLIP-T, verifying that Cross-Contextual Bridging effectively enhances text
\begin{wrapfigure}[12]{r}{0.5\textwidth}
\vspace{-10pt}
\centering
\includegraphics[width=0.49\textwidth]{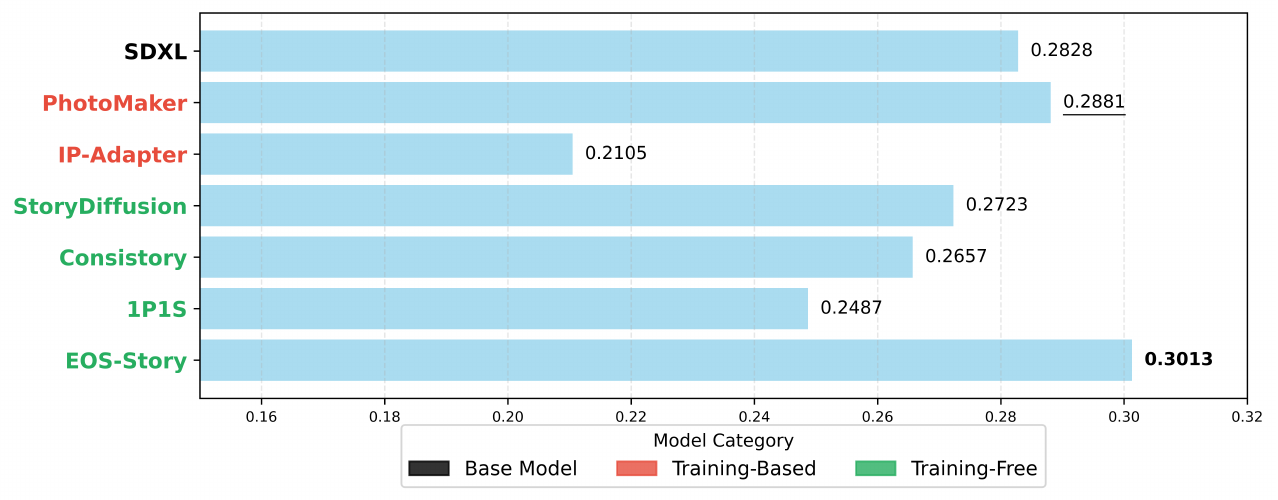}
\caption{
\textbf{Top-1 framing classification accuracy on Consistory+\_+.} Evaluated with ShotVL-7B~\cite{liu2025shotbench}, EOS-Story effectively captures the intended framing phrases.
}
\label{fig:shot_acc}
\end{wrapfigure}
alignment.
While CLIP-I is slightly below the best, it remains competitive with the highest score and achieves the best DreamSim, demonstrating effective identity preservation.
We also observe substantial improvements in image quality metrics (Aesthetic, ImageReward), confirming EOS-Story’s ability to maintain visual fidelity.
For framing accuracy (\cref{fig:shot_acc}), which evaluates whether the framing phrases are properly reflected, EOS-Story also achieves the best results on Consistory+\_+.
These findings further corroborate the effectiveness of our method for adaptively capturing scene variances.

\begin{figure*}[!t]
    \centering
    \includegraphics[width=0.97\textwidth]{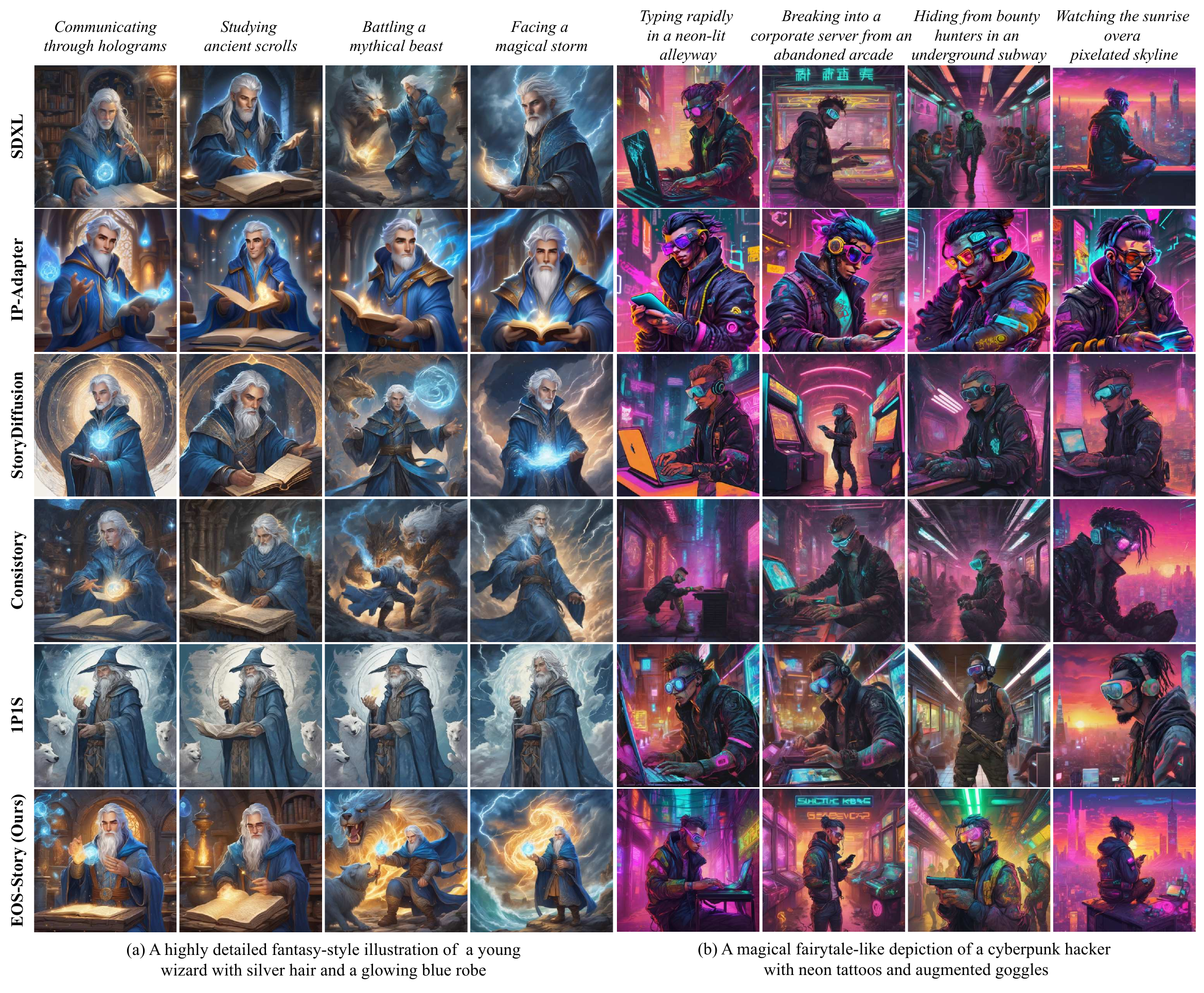}
    \caption{\textbf{Image comparison with previous methods.} Our method faithfully reflects the given scenario while showing a flexible subject in each image.}
    \label{fig:quality1}
\end{figure*}

\begin{figure*}[!t]
\centering
\begin{minipage}{0.48\textwidth}
    \centering
    \includegraphics[width=\textwidth]{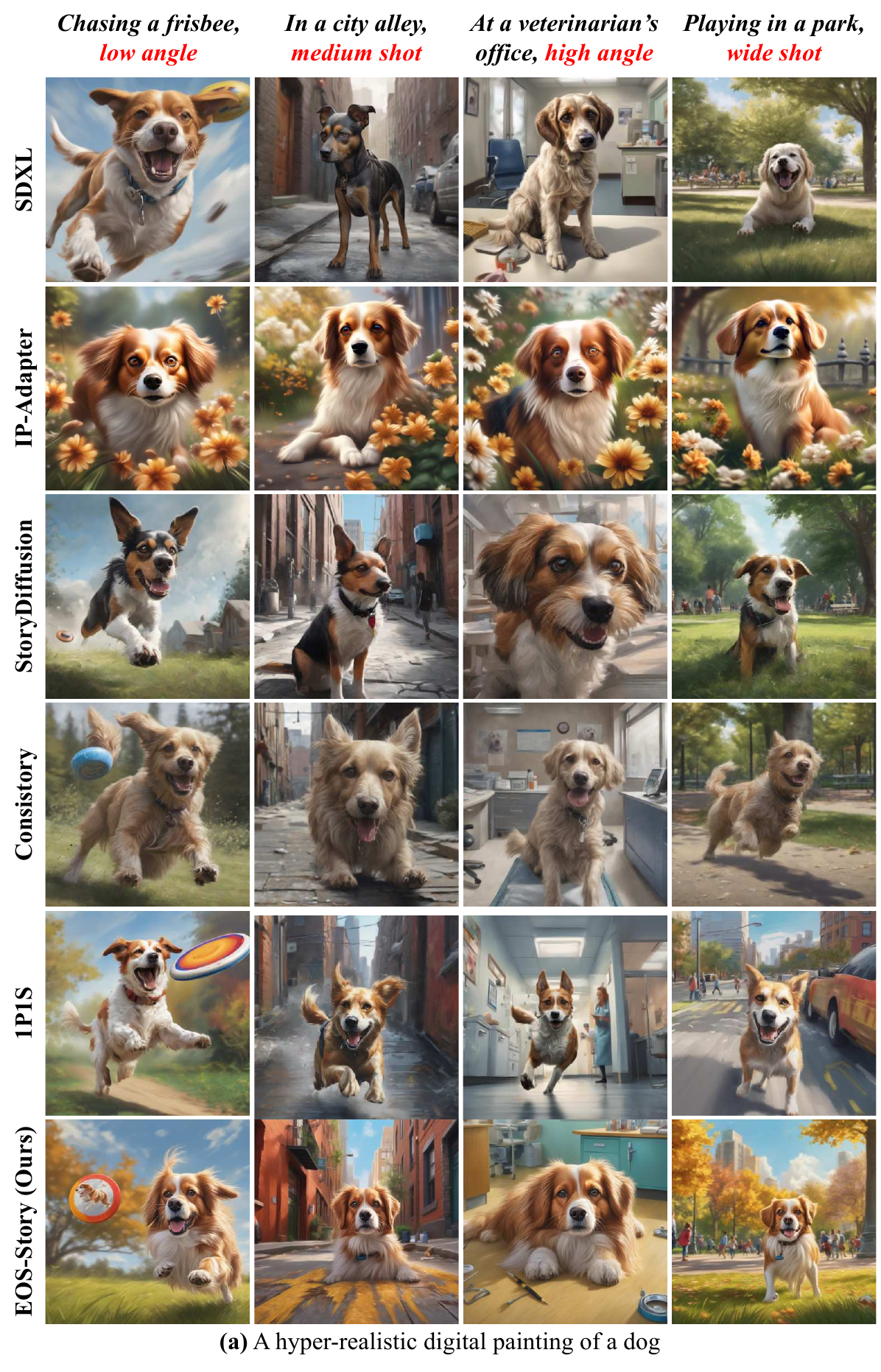}
    \caption{\textbf{Image difference with various framing phrases.} 
    We show a line of storyboard from our Consistory+\_+.}
    \label{fig:quality2}
\end{minipage}
\hfill
\begin{minipage}{0.48\textwidth}
    \centering
    \includegraphics[width=\textwidth]{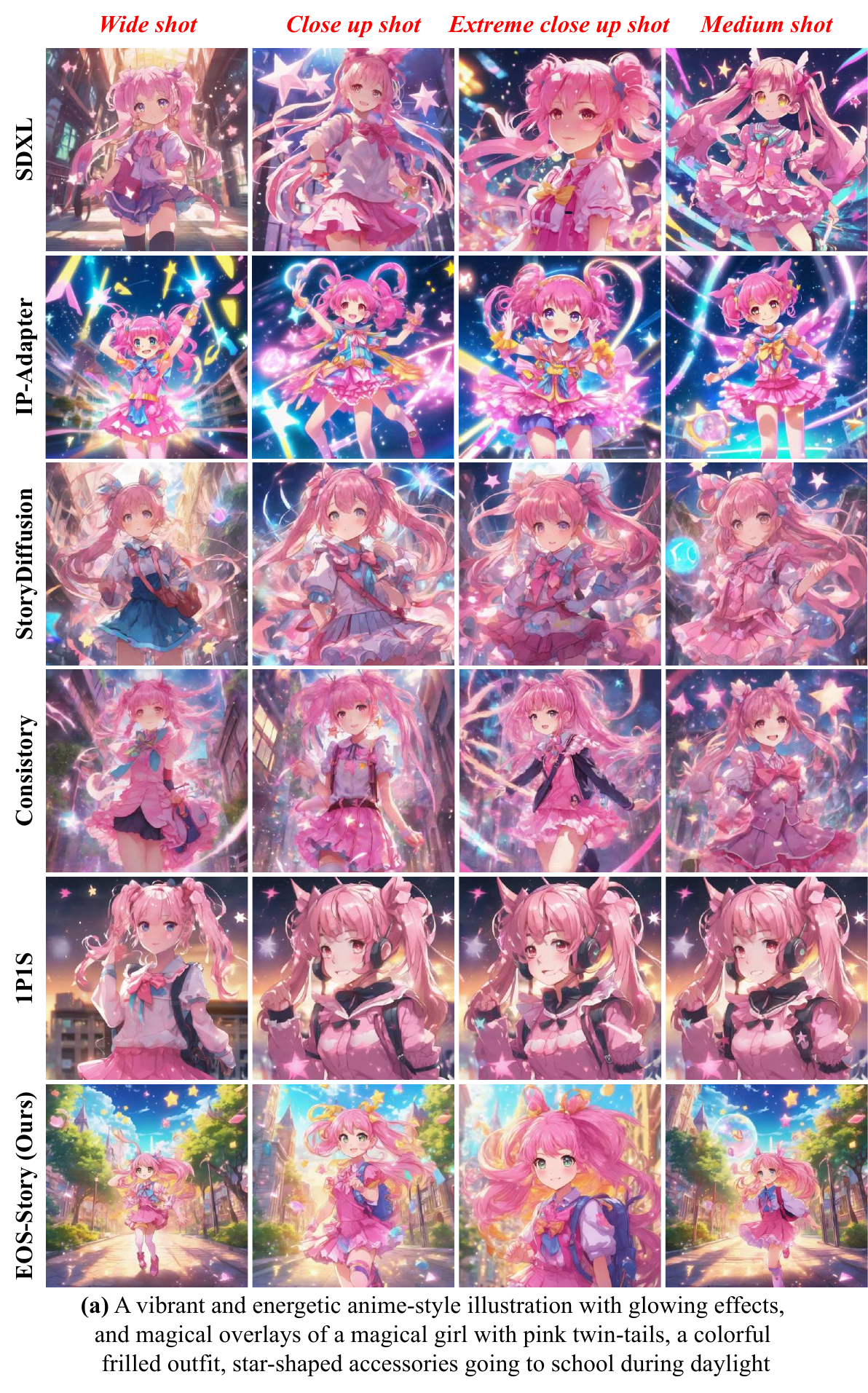}
    \caption{\textbf{Viewpoint adaptation comparison.} 
    Various viewpoints are appended to the end of the same scene.}
    \label{fig:quality3}
\end{minipage}
\end{figure*}

\noindent\textbf{Qualitative Result.} We first compare the results in \cref{fig:quality1}.
Across diverse scenes, our model generates images that are both textually and visually well-aligned compared to prior methods.
For instance, some models produce structurally fixed or repetitive outputs across varied scenes (\cref{fig:quality1}a row 5), while others fail to reflect key attributes (e.g., underrepresenting subject details such as the ``beard'' of wizard in \cref{fig:quality1}a).
On Consistory+\_+, which includes various framing phrases (\cref{fig:quality2}), our model successfully captures the intended framing while preserving both prompt and subject fidelity across scenes.
Interestingly, under scenarios with extremely long prompts (\cref{fig:quality3}), our model robustly adapts framing structures according to the various viewpoints, whereas prior methods often struggle to do so.
These results demonstrate that EOS-Story effectively adapts to scene variance, ultimately producing enhanced visual storytelling outputs.
Additional comparisons and results are in the supplementary materials.



\subsection{Effectiveness of Cross-Contextual Bridging}
\label{eos_analysis}
Building upon the empirical observation in \cref{sec:eos_emp} that the $\langle$EOS$\rangle$ embedding acts as a global semantic control point, we conduct a quantitative evaluation in addition to providing qualitative samples.
As shown in \cref{tab:token_ablation}, injecting appearance information solely into the subject token fails to reliably encode identity cues.
Distributing the same information across all non-EOS tokens partially improves alignment, yet the gains remain limited, suggesting that localized token-level injection struggles to propagate appearance semantics consistently during generation.
In contrast, modifying only the $\langle$EOS$\rangle$ embedding consistently achieves the highest text–image alignment and perceptual similarity. 
These results indicate that semantic control is more effectively exercised at the EOS level, where global context is integrated, rather than at individual word tokens that contribute only localized representations.

To further analyze the design choice of our residual formulation, we compare $eos_{inf}$ with $eos_{org}$, where $eos_{org}$ is obtained by holistically encoding the full original prompt (\textcolor{style}{style} + subject + \textcolor{appearance}{appearance} + \textcolor{scene}{scene}). 
We further consider two alternative constructions: (1) directly adding $eos_{id}$ to $eos_{sc}$, and (2) an orthogonal-based method following~\cite{dalva2024fluxspace}, where we construct an orthogonalized appearance component $\Delta_{app}^{orth} = eos_{id} - \frac{eos_{id} \cdot eos_{sty}}{||eos_{sty}||^2} eos_{sty}$.
We begin by constructing two semantic subspaces using PCA, based on the $\langle$EOS$\rangle$ embeddings obtained from appearance and scene phrases in the Consistory+ benchmark.

To ensure reliability, we exclude samples lacking explicit appearance descriptions (e.g., prompts specifying only a subject without appearance descriptors). 
Next, we project each EOS embedding onto these subspaces and compute the average cosine similarity to quantify alignment with each semantic axis.
\begin{figure*}[t]
\centering

\begin{minipage}[t]{0.46\textwidth}
\vspace{0pt}
\centering

\captionof{table}{\textbf{Various position for injecting information.}
Injecting solely on EOS token shows the best score.}
\label{tab:token_ablation}
\vspace{1mm}

\scriptsize
\setlength{\tabcolsep}{3.5pt}
\renewcommand{\arraystretch}{1.2}

\resizebox{\linewidth}{!}{
\begin{tabular}{l|ccc}
\hline \hline
& CLIP-T$\uparrow$ & CLIP-I$\uparrow$ & DreamSim$\downarrow$ \\ \hline
Baseline (SDXL) & 0.8971 & 0.8430 & 0.3124 \\ 
Subject token & 0.8909 & 0.8782 & 0.2341 \\
All w/o EOS   & \underline{0.8990} & \underline{0.8823} & \underline{0.2150} \\ \hline
Only EOS (Ours) & \textbf{0.9042} & \textbf{0.8850} & \textbf{0.2005} \\
\hline \hline
\end{tabular}
}
\end{minipage}
\hfill
\begin{minipage}[t]{0.52\textwidth}
\vspace{0pt}
\centering

\captionof{table}{\textbf{Subspace alignment of $eos_{inf}$.}
Alignment scores on appearance/scene subspaces and CLIP-T for selected EOS variants.}
\label{tab:subspace_alignment}
\vspace{1mm}

\footnotesize
\renewcommand{\arraystretch}{0.95}
\setlength{\tabcolsep}{6pt}

\resizebox{\linewidth}{!}{
\begin{tabular}{l|ccc}
\hline \hline
 & Appearance$\uparrow$ & Scene$\uparrow$ & CLIP-T$\uparrow$ \\ \hline
(1) $eos_{org}$  & \underline{0.555} & 0.692 & 0.8935 \\
(2) $eos_{sc} + eos_{id}$ & 0.519 & 0.655 & 0.8925 \\
(3) $eos_{sc} + \Delta_{app}^{orth}$ & 0.552 & \underline{0.694} & \underline{0.9033} \\ \hline
$eos_{sc} + \Delta_{app}$ ($eos_{inf}$) & \textbf{0.564} & \textbf{0.706} & \textbf{0.9042} \\
\hline \hline
\end{tabular}
}

\end{minipage}

\vspace{-2mm}
\end{figure*}
As shown in \cref{tab:subspace_alignment}, directly adding $eos_{id}$ degrades alignment, likely due to redundant semantic components that disrupt scene coherence. 
The orthogonal variant partially alleviates this issue, but still falls short of containing both appearance and scene semantics. 
In contrast, our residual formulation ($eos_{sc} + \Delta_{app}$) achieves the highest alignment across both semantic subspaces and CLIP-T.
Taken together,
\begin{wrapfigure}[12]{r}{0.48\textwidth}
\vspace{-4mm}
\centering
\includegraphics[width=\linewidth]{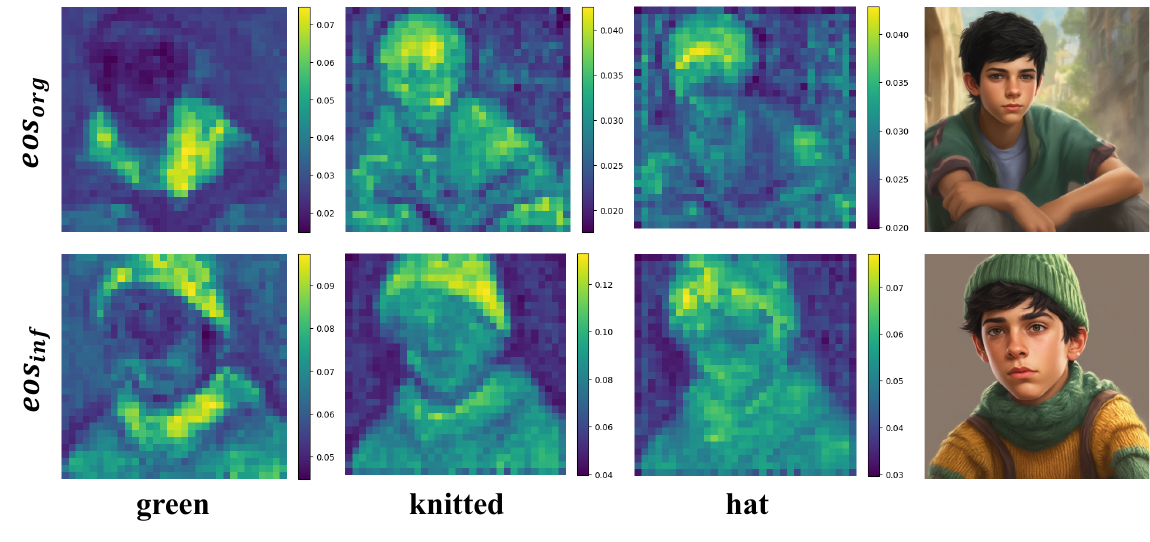}
\caption{\textbf{Attention sample visualization.}
$eos_{inf}$ attends more strongly to semantically relevant regions.}
\label{fig:linear4}
\vspace{-4mm}
\end{wrapfigure}
these results indicate that residual appearance injection at the EOS token helps effectively reflect the intended semantics during generation.
To qualitatively support this analysis, \cref{fig:linear4} visualizes cross-attention maps. 
Compared to $eos_{org}$, which exhibits dispersed attention over irrelevant regions, 
$eos_{inf}$ concentrates attention on semantically relevant areas, such as correctly attending to the specific details (e.g., ``green knitted hat''). 
This focused attention pattern suggests that our EOS residual formulation helps mitigate semantic dilution, enabling the model to selectively attend to scene-specific cues during generation.
Additional analyses are in the supplementary materials.

\subsection{Ablation Studies}

\begin{figure*}[!t]
\centering
\begin{minipage}[t]{0.46\textwidth}
    \centering
    \vspace{0pt}
    \includegraphics[width=\linewidth]{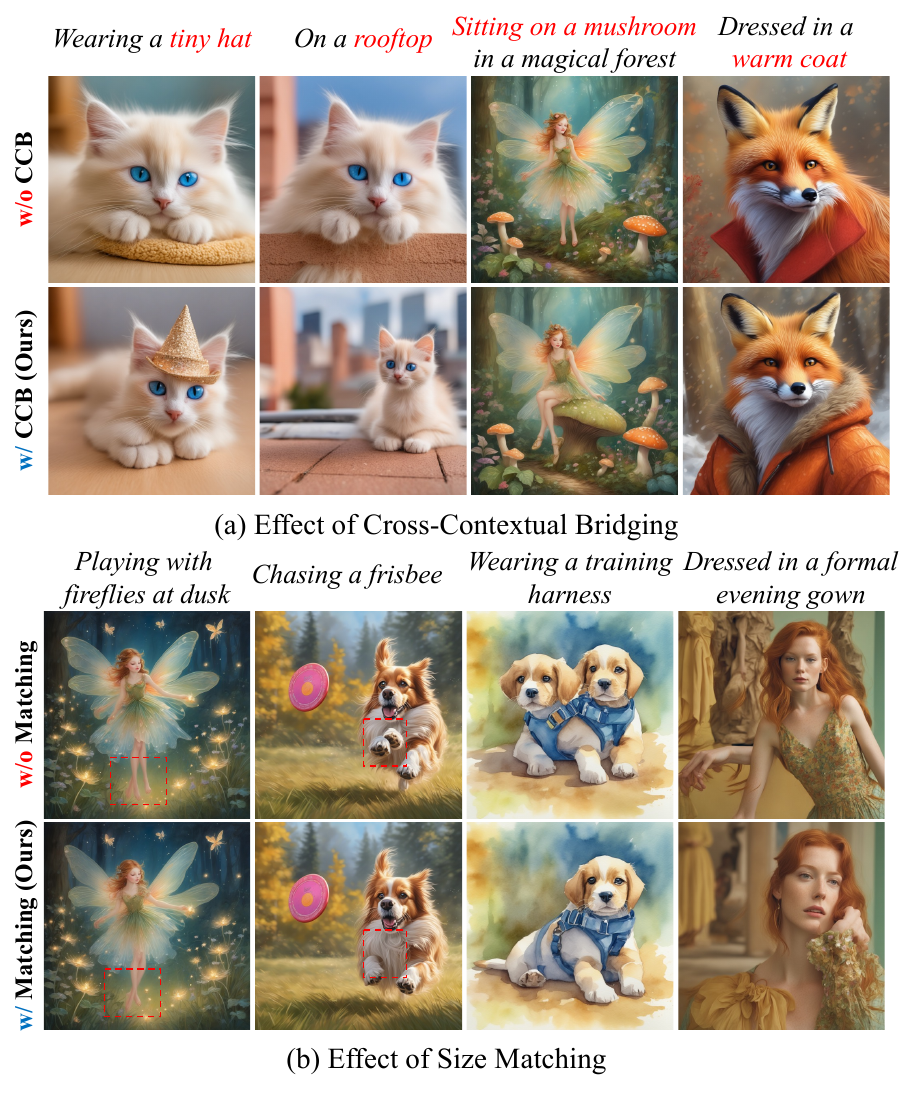}
    \captionof{figure}{\textbf{Component ablation results.}
    CCB improves alignment, while removing size matching causes structural errors.}
    \label{fig:image_abla}
\end{minipage}
\hfill
\begin{minipage}[t]{0.52\textwidth}
    \centering
    \vspace{0pt}
    \captionof{table}{\textbf{Score difference across components.} 
    Combining all components effectively balances subject and prompt coherence.}
    \label{tab:component}
    \vspace{1mm}
    \footnotesize
    \setlength{\tabcolsep}{3pt}
    \renewcommand{\arraystretch}{0.95}
    \resizebox{\linewidth}{!}{%
    \begin{tabular}{c|ccc|ccc}
    \hline \hline
    ~ & CCB & IA & IASA & CLIP-T & CLIP-I & DreamSim \\ \hline
    SDXL &  &  &  & 0.8971 & 0.8430 & 0.3124 \\
    (2)  & \textcolor{green!60!black}{\ding{51}} &  &  & 0.8974 & 0.8290 & 0.3067 \\
    (3)  &  & \textcolor{green!60!black}{\ding{51}} &  & 0.8924 & 0.8729 & 0.2348 \\
    (4)  &  &  & \textcolor{green!60!black}{\ding{51}} & 0.8924 & 0.8730 & 0.2348 \\
    (5)  & \textcolor{green!60!black}{\ding{51}} &  & \textcolor{green!60!black}{\ding{51}} & \underline{0.9037} & 0.8743 & 0.2164 \\
    (6)  & \textcolor{green!60!black}{\ding{51}} & \textcolor{green!60!black}{\ding{51}} &  & \underline{0.9037} & 0.8742 & 0.2167 \\
    (7)  &  & \textcolor{green!60!black}{\ding{51}} & \textcolor{green!60!black}{\ding{51}} & 0.8935 & \textbf{0.8924} & \underline{0.2105} \\ \hline
    Ours & \textcolor{green!60!black}{\ding{51}} & \textcolor{green!60!black}{\ding{51}} & \textcolor{green!60!black}{\ding{51}} & \textbf{0.9042} & \underline{0.8850} & \textbf{0.2005} \\
    \hline \hline
    \end{tabular}%
    }

    \vspace{4mm}

    \captionof{table}{\textbf{Effect of size matching.} 
    Size matching promotes structural coherence and improves image quality.}
    \label{tab:size_match_abla}
    \vspace{1mm}
    \footnotesize
    \renewcommand{\arraystretch}{0.95}
    \setlength{\tabcolsep}{4pt}
    \resizebox{\linewidth}{!}{%
    \begin{tabular}{l|ccc}
    \hline \hline
     & CLIP-I & Aesthetic & ImageReward \\ \hline
    \textcolor{red}{w/o} match & 0.8842 & 6.1749 & 1.1618 \\
    \textcolor{blue}{w/} match & \textbf{0.8850} & \textbf{6.1764} & \textbf{1.1680} \\
    \hline \hline
    \end{tabular}%
    }
\end{minipage}
\end{figure*}

We conduct ablation studies on Consistory+. Additional experimental results are provided in the supplementary material.

\noindent\textbf{Component Ablation.}
In \cref{tab:component}, we report score variations across different component configurations. 
Applying Cross-Contextual Bridging (CCB) consistently improves CLIP-T, demonstrating enhanced text alignment.
This is further supported by \cref{fig:image_abla}a, where intended actions, objects, or situations are more accurately depicted.
Notably, improvements from CCB are primarily reflected in text-related metrics, whereas Adaptive Identity Preservation contributes more prominently to identity-related scores such as CLIP-I. 
This complementary trend indicates that each module effectively targets its intended objective, enabling more balanced control over textual fidelity and identity consistency.
While configuration (7) achieves higher CLIP-I, our full method, EOS-Story, outperforms in both CLIP-T and DreamSim.
We also compare results with and without subject size matching (\cref{fig:image_abla}b, \cref{tab:size_match_abla}). 
Without size aligning, scale discrepancies between identity and scene representations often lead to misaligned identity transfer, resulting in distorted subject proportions. 
In contrast, size-aware alignment ensures that identity features are referenced at an appropriate spatial scale, promoting more coherent subject integration.
\section{Conclusion}
In this work, we propose \textbf{EOS-Story}, a training-free framework that adaptively models scene-specific variance in visual storytelling across text and vision spaces. We introduce \textbf{Cross-Contextual Bridging}, which decouples the narrative into identity and scene sub-prompts and bridges the resulting semantic gap by injecting phrase-level appearance residuals into the scene's $\langle$EOS$\rangle$ embedding. To complement this with structural adaptation, we further propose \textbf{Adaptive Identity Preservation}, which anchors subject features at the scene-specific scale and refines integration through Identity-Aware Self-Attention. To evaluate scene adaptability, we extend Consistory+ with diverse framing variations (Consistory+\_+).
Extensive experiments show that EOS-Story achieves competitive or superior performance in both prompt fidelity and structural adaptability.

\par\vspace{\baselineskip}
\noindent\textbf{Acknowledgements.}
This work was partly supported by the Institute of Information \&
Communications Technology Planning \& Evaluation (IITP) grant funded by
the Korean government (MSIT) (No. RS-2020-II201373, Artificial Intelligence
Graduate School Program (Hanyang University) and No. RS-2025-02219062,
Self-training framework for VLM-based defect detection and explanation
model in manufacturing process).

%
%
\bibliographystyle{splncs04}
\bibliography{main}

@String(ICCV  = {Int. Conf. Comput. Vis.})

@String(TOG   = {ACM Trans. Graph.})

@String(ICCV  = {ICCV})

@String(TOG   = {ACM TOG})

@article{liu2025one,
  title={One-prompt-one-story: Free-lunch consistent text-to-image generation using a single prompt},
  author={Liu, Tao and Wang, Kai and Li, Senmao and van de Weijer, Joost and Khan, Fahad Shahbaz and Yang, Shiqi and Wang, Yaxing and Yang, Jian and Cheng, Ming-Ming},
  journal={arXiv preprint arXiv:2501.13554},
  year={2025}
}

@article{zhou2024storydiffusion,
  title={Storydiffusion: Consistent self-attention for long-range image and video generation},
  author={Zhou, Yupeng and Zhou, Daquan and Cheng, Ming-Ming and Feng, Jiashi and Hou, Qibin},
  journal={Advances in Neural Information Processing Systems},
  volume={37},
  pages={110315--110340},
  year={2024}
}

@article{tewel2024training,
  title={Training-free consistent text-to-image generation},
  author={Tewel, Yoad and Kaduri, Omri and Gal, Rinon and Kasten, Yoni and Wolf, Lior and Chechik, Gal and Atzmon, Yuval},
  journal={ACM Transactions on Graphics (TOG)},
  volume={43},
  number={4},
  pages={1--18},
  year={2024},
  publisher={ACM New York, NY, USA}
}

@article{podell2023sdxl,
  title={Sdxl: Improving latent diffusion models for high-resolution image synthesis},
  author={Podell, Dustin and English, Zion and Lacey, Kyle and Blattmann, Andreas and Dockhorn, Tim and M{\"u}ller, Jonas and Penna, Joe and Rombach, Robin},
  journal={arXiv preprint arXiv:2307.01952},
  year={2023}
}

@inproceedings{rombach2022high,
  title={High-resolution image synthesis with latent diffusion models},
  author={Rombach, Robin and Blattmann, Andreas and Lorenz, Dominik and Esser, Patrick and Ommer, Bj{\"o}rn},
  booktitle={Proceedings of the IEEE/CVF conference on computer vision and pattern recognition},
  pages={10684--10695},
  year={2022}
}

@article{yi2024towards,
  title={Towards understanding the working mechanism of text-to-image diffusion model},
  author={Yi, Mingyang and Li, Aoxue and Xin, Yi and Li, Zhenguo},
  journal={Advances in Neural Information Processing Systems},
  volume={37},
  pages={55342--55369},
  year={2024}
}

@article{mao2024story,
  title={Story-adapter: A training-free iterative framework for long story visualization},
  author={Mao, Jiawei and Huang, Xiaoke and Xie, Yunfei and Chang, Yuanqi and Hui, Mude and Xu, Bingjie and Zhou, Yuyin},
  journal={arXiv preprint arXiv:2410.06244},
  year={2024}
}

@article{wang2024oneactor,
  title={OneActor: Consistent Subject Generation via Cluster-Conditioned Guidance},
  author={Wang, Jiahao and Yan, Caixia and Lin, Haonan and Zhang, Weizhan and Wang, Mengmeng and Gong, Tieliang and Dai, Guang and Sun, Hao},
  journal={Advances in Neural Information Processing Systems},
  volume={37},
  pages={21502--21536},
  year={2024}
}

@inproceedings{rahman2023make,
  title={Make-a-story: Visual memory conditioned consistent story generation},
  author={Rahman, Tanzila and Lee, Hsin-Ying and Ren, Jian and Tulyakov, Sergey and Mahajan, Shweta and Sigal, Leonid},
  booktitle={Proceedings of the IEEE/CVF conference on computer vision and pattern recognition},
  pages={2493--2502},
  year={2023}
}

@inproceedings{radford2021learning,
  title={Learning transferable visual models from natural language supervision},
  author={Radford, Alec and Kim, Jong Wook and Hallacy, Chris and Ramesh, Aditya and Goh, Gabriel and Agarwal, Sandhini and Sastry, Girish and Askell, Amanda and Mishkin, Pamela and Clark, Jack and others},
  booktitle={International conference on machine learning},
  pages={8748--8763},
  year={2021},
  organization={PmLR}
}

@article{mun2024addressing,
  title={Addressing Attribute Leakages in Diffusion-based Image Editing without Training},
  author={Mun, Sunung and Nam, Jinhwan and Cho, Sunghyun and Ok, Jungseul},
  journal={arXiv preprint arXiv:2412.04715},
  year={2024}
}

@article{hu2024token,
  title={Token merging for training-free semantic binding in text-to-image synthesis},
  author={Hu, Taihang and Li, Linxuan and van de Weijer, Joost and Gao, Hongcheng and Shahbaz Khan, Fahad and Yang, Jian and Cheng, Ming-Ming and Wang, Kai and Wang, Yaxing},
  journal={Advances in Neural Information Processing Systems},
  volume={37},
  pages={137646--137672},
  year={2024}
}

@article{tang2023emergent,
  title={Emergent correspondence from image diffusion},
  author={Tang, Luming and Jia, Menglin and Wang, Qianqian and Phoo, Cheng Perng and Hariharan, Bharath},
  journal={Advances in Neural Information Processing Systems},
  volume={36},
  pages={1363--1389},
  year={2023}
}

@article{otsu1975threshold,
  title={A threshold selection method from gray-level histograms},
  author={Otsu, Nobuyuki and others},
  journal={Automatica},
  volume={11},
  number={285-296},
  pages={23--27},
  year={1975}
}

@article{phung2025cineverse,
  title={CineVerse: Consistent Keyframe Synthesis for Cinematic Scene Composition},
  author={Phung, Quynh and Mai, Long and Heilbron, Fabian David Caba and Liu, Feng and Huang, Jia-Bin and Ham, Cusuh},
  journal={arXiv preprint arXiv:2504.19894},
  year={2025}
}

@inproceedings{trager2023linear,
  title={Linear spaces of meanings: compositional structures in vision-language models},
  author={Trager, Matthew and Perera, Pramuditha and Zancato, Luca and Achille, Alessandro and Bhatia, Parminder and Soatto, Stefano},
  booktitle={Proceedings of the IEEE/CVF International Conference on Computer Vision},
  pages={15395--15404},
  year={2023}
}

@article{liu2024improving,
  title={Improving long-text alignment for text-to-image diffusion models},
  author={Liu, Luping and Du, Chao and Pang, Tianyu and Wang, Zehan and Li, Chongxuan and Xu, Dong},
  journal={arXiv preprint arXiv:2410.11817},
  year={2024}
}

@article{liu2025shotbench,
  title={ShotBench: Expert-Level Cinematic Understanding in Vision-Language Models},
  author={Liu, Hongbo and He, Jingwen and Jin, Yi and Zheng, Dian and Dong, Yuhao and Zhang, Fan and Huang, Ziqi and He, Yinan and Li, Yangguang and Chen, Weichao and others},
  journal={arXiv preprint arXiv:2506.21356},
  year={2025}
}

@article{ye2023ip,
  title={Ip-adapter: Text compatible image prompt adapter for text-to-image diffusion models},
  author={Ye, Hu and Zhang, Jun and Liu, Sibo and Han, Xiao and Yang, Wei},
  journal={arXiv preprint arXiv:2308.06721},
  year={2023}
}

@inproceedings{li2024photomaker,
  title={Photomaker: Customizing realistic human photos via stacked id embedding},
  author={Li, Zhen and Cao, Mingdeng and Wang, Xintao and Qi, Zhongang and Cheng, Ming-Ming and Shan, Ying},
  booktitle={Proceedings of the IEEE/CVF conference on computer vision and pattern recognition},
  pages={8640--8650},
  year={2024}
}

@inproceedings{avrahami2024chosen,
  title={The chosen one: Consistent characters in text-to-image diffusion models},
  author={Avrahami, Omri and Hertz, Amir and Vinker, Yael and Arar, Moab and Fruchter, Shlomi and Fried, Ohad and Cohen-Or, Daniel and Lischinski, Dani},
  booktitle={ACM SIGGRAPH 2024 conference papers},
  pages={1--12},
  year={2024}
}

@article{gal2022image,
  title={An image is worth one word: Personalizing text-to-image generation using textual inversion},
  author={Gal, Rinon and Alaluf, Yuval and Atzmon, Yuval and Patashnik, Or and Bermano, Amit H and Chechik, Gal and Cohen-Or, Daniel},
  journal={arXiv preprint arXiv:2208.01618},
  year={2022}
}

@inproceedings{ruiz2023dreambooth,
  title={Dreambooth: Fine tuning text-to-image diffusion models for subject-driven generation},
  author={Ruiz, Nataniel and Li, Yuanzhen and Jampani, Varun and Pritch, Yael and Rubinstein, Michael and Aberman, Kfir},
  booktitle={Proceedings of the IEEE/CVF conference on computer vision and pattern recognition},
  pages={22500--22510},
  year={2023}
}

@article{hurst2024gpt,
  title={Gpt-4o system card},
  author={Hurst, Aaron and Lerer, Adam and Goucher, Adam P and Perelman, Adam and Ramesh, Aditya and Clark, Aidan and Ostrow, AJ and Welihinda, Akila and Hayes, Alan and Radford, Alec and others},
  journal={arXiv preprint arXiv:2410.21276},
  year={2024}
}

@article{gao2025subject,
  title={Subject-Consistent and Pose-Diverse Text-to-Image Generation},
  author={Gao, Zhanxin and Zhu, Beier and Yao, Liang and Yang, Jian and Tai, Ying},
  journal={arXiv preprint arXiv:2507.08396},
  year={2025}
}

@article{li2023blip,
  title={Blip-diffusion: Pre-trained subject representation for controllable text-to-image generation and editing},
  author={Li, Dongxu and Li, Junnan and Hoi, Steven},
  journal={Advances in Neural Information Processing Systems},
  volume={36},
  pages={30146--30166},
  year={2023}
}

@misc{taited2023CLIPScore,
  author={SUN Zhengwentai},
  title={{clip-score: CLIP Score for PyTorch}},
  month={March},
  year={2023},
  note={Version 0.2.1},
  howpublished={\url{https://github.com/taited/clip-score}},
}

@article{fu2023dreamsim,
  title={Dreamsim: Learning new dimensions of human visual similarity using synthetic data},
  author={Fu, Stephanie and Tamir, Netanel and Sundaram, Shobhita and Chai, Lucy and Zhang, Richard and Dekel, Tali and Isola, Phillip},
  journal={arXiv preprint arXiv:2306.09344},
  year={2023}
}

@article{xu2023imagereward,
  title={Imagereward: Learning and evaluating human preferences for text-to-image generation},
  author={Xu, Jiazheng and Liu, Xiao and Wu, Yuchen and Tong, Yuxuan and Li, Qinkai and Ding, Ming and Tang, Jie and Dong, Yuxiao},
  journal={Advances in Neural Information Processing Systems},
  volume={36},
  pages={15903--15935},
  year={2023}
}

@misc{discus2024aestheticv25,
  author       = {{discus0434}},
  title        = {Aesthetic Predictor v2.5},
  howpublished = {\url{https://github.com/discus0434/aesthetic-predictor-v2-5}},
  year         = {2024},
  note         = {GitHub repository, accessed 2025-09-28}
}

@inproceedings{zhai2023sigmoid,
  title={Sigmoid loss for language image pre-training},
  author={Zhai, Xiaohua and Mustafa, Basil and Kolesnikov, Alexander and Beyer, Lucas},
  booktitle={Proceedings of the IEEE/CVF international conference on computer vision},
  pages={11975--11986},
  year={2023}
}

@article{dalva2024fluxspace,
  title={Fluxspace: Disentangled semantic editing in rectified flow transformers},
  author={Dalva, Yusuf and Venkatesh, Kavana and Yanardag, Pinar},
  journal={arXiv preprint arXiv:2412.09611},
  year={2024}
}

@article{lian2023llm,
  title={Llm-grounded diffusion: Enhancing prompt understanding of text-to-image diffusion models with large language models},
  author={Lian, Long and Li, Boyi and Yala, Adam and Darrell, Trevor},
  journal={arXiv preprint arXiv:2305.13655},
  year={2023}
}

@article{Rooney2018WatchingMC,
  title={Watching More Closely: Shot Scale Affects Film Viewers’ Theory of Mind Tendency But Not Ability},
  author={Brendan Rooney and Katalin E. B{\'a}lint},
  journal={Frontiers in Psychology},
  year={2018},
  volume={8},
  url={https://api.semanticscholar.org/CorpusID:3754212}
}

@InProceedings{Wang_2025_ICCV,
    author    = {Wang, Mengyu and Ding, Henghui and Peng, Jianing and Zhao, Yao and Chen, Yunpeng and Wei, Yunchao},
    title     = {CharaConsist: Fine-Grained Consistent Character Generation},
    booktitle = {Proceedings of the IEEE/CVF International Conference on Computer Vision (ICCV)},
    month     = {October},
    year      = {2025},
    pages     = {16058-16067}
}

@inproceedings{esser2024scaling,
  title={Scaling rectified flow transformers for high-resolution image synthesis},
  author={Esser, Patrick and Kulal, Sumith and Blattmann, Andreas and Entezari, Rahim and M{\"u}ller, Jonas and Saini, Harry and Levi, Yam and Lorenz, Dominik and Sauer, Axel and Boesel, Frederic and others},
  booktitle={Forty-first international conference on machine learning},
  year={2024}
}

@article{ding2024longrope,
  title={Longrope: Extending llm context window beyond 2 million tokens},
  author={Ding, Yiran and Zhang, Li Lyna and Zhang, Chengruidong and Xu, Yuanyuan and Shang, Ning and Xu, Jiahang and Yang, Fan and Yang, Mao},
  journal={arXiv preprint arXiv:2402.13753},
  year={2024}
}

@article{lu2024fit,
  title={Fit: Flexible vision transformer for diffusion model},
  author={Lu, Zeyu and Wang, Zidong and Huang, Di and Wu, Chengyue and Liu, Xihui and Ouyang, Wanli and Bai, Lei},
  journal={arXiv preprint arXiv:2402.12376},
  year={2024}
}

@inproceedings{toker2025padding,
  title={Padding tone: A mechanistic analysis of padding tokens in t2i models},
  author={Toker, Michael and Galil, Ido and Orgad, Hadas and Gal, Rinon and Tewel, Yoad and Chechik, Gal and Belinkov, Yonatan},
  booktitle={Proceedings of the 2025 Conference of the Nations of the Americas Chapter of the Association for Computational Linguistics: Human Language Technologies (Volume 1: Long Papers)},
  pages={7618--7632},
  year={2025}
}

@inproceedings{li2024get,
  title={Get what you want, not what you don't: Image content suppression for text-to-image diffusion models},
  author={Li, Senmao and Van de Weijer, Joost and Khan, Fahad and Hou, Qibin and Wang, Yaxing and others},
  booktitle={International Conference on Learning Representations},
  volume={2024},
  pages={3182--3208},
  year={2024}
}

\clearpage
\appendix
\begin{center}
{\LARGE\bfseries Supplementary Materials}\\[0.5em]
{\Large\bfseries Not Just a Subject: Capturing Adaptive Scene-Specific Variance for Visual Storytelling}\\[0.25em]
\end{center}
\vspace{1em}
\FloatBarrier

\begin{figure*}[!h]
    \centering
    \includegraphics[width=\textwidth]{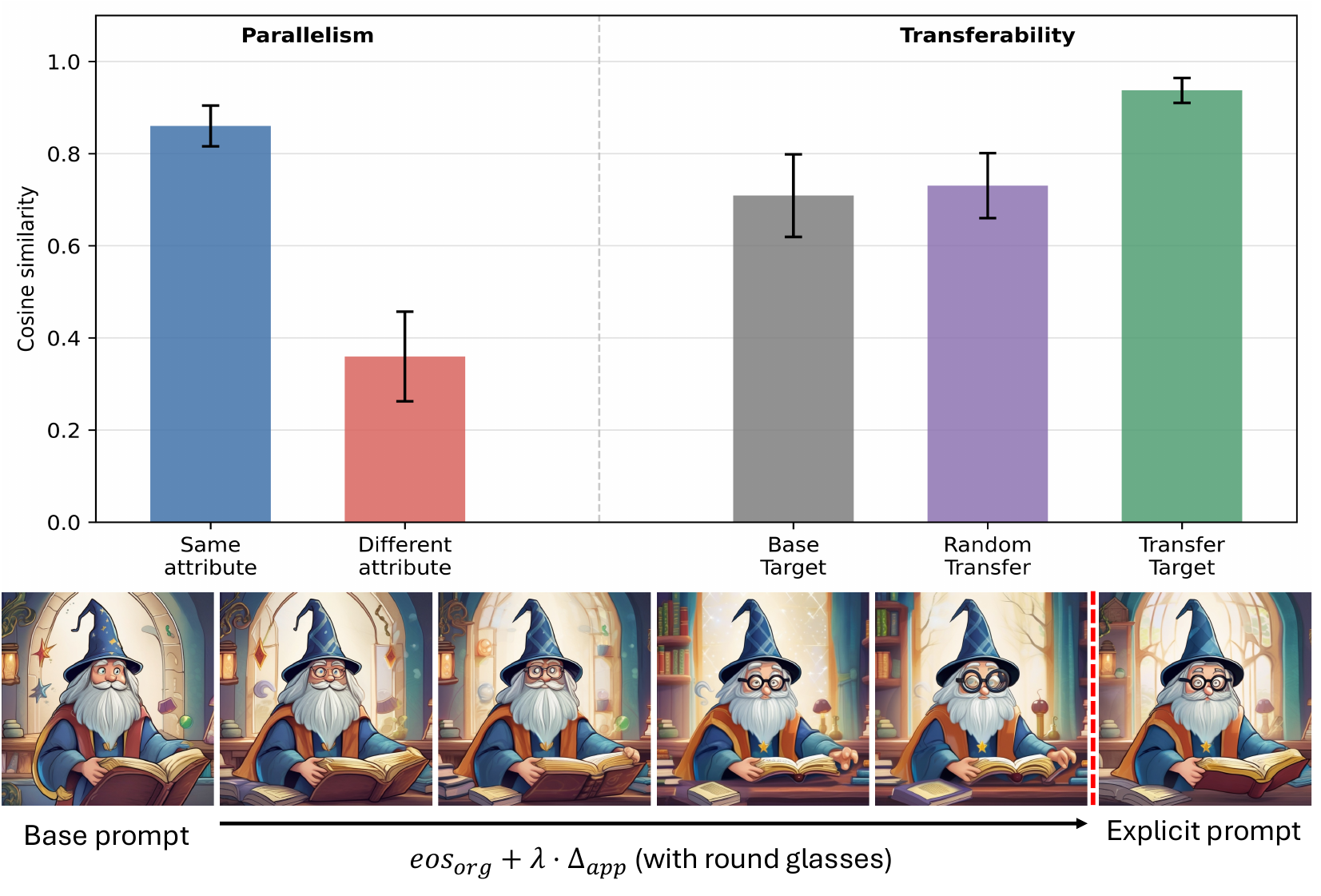}
\captionof{figure}{\textbf{Empirical analysis of ⟨EOS⟩ linearity.} \textbf{(Parallelism)} $\Delta_{app}$ from the same attribute stays aligned across subjects but separates across attributes, indicating a subject-agnostic direction. \textbf{(Transferability)} A transferred residual matches the explicit-prompt embedding (\textit{transfer target}), unlike the base or a random attribute. \textbf{(Bottom)} Increasing $\lambda$ in progressively strengthens the target appearance (round glasses), confirming controllable influence.}
    \label{fig:eos_linearity}
\end{figure*}

\begin{figure}[!h]
    \centering
    \begin{minipage}[t]{0.49\textwidth}
        \centering
            \includegraphics[width=0.9\linewidth]{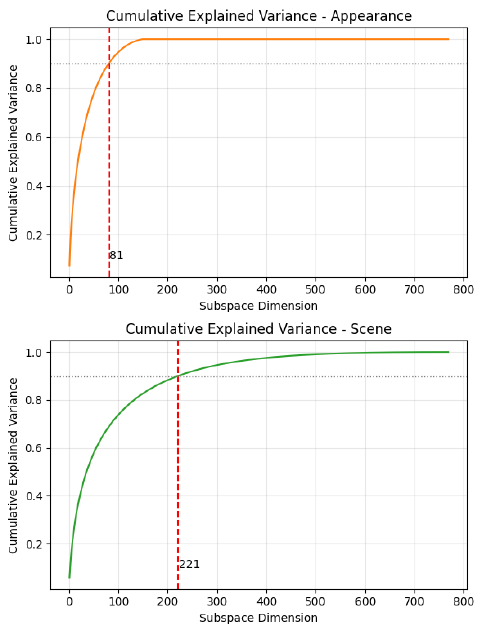}
        \captionof{figure}{
        \textbf{Cumulative explained variance for scene and appearance subspaces.}}
        \label{fig:pca_cv}
    \end{minipage}
    \hfill
    \begin{minipage}[t]{0.49\textwidth}
        \centering
        \includegraphics[width=\linewidth]{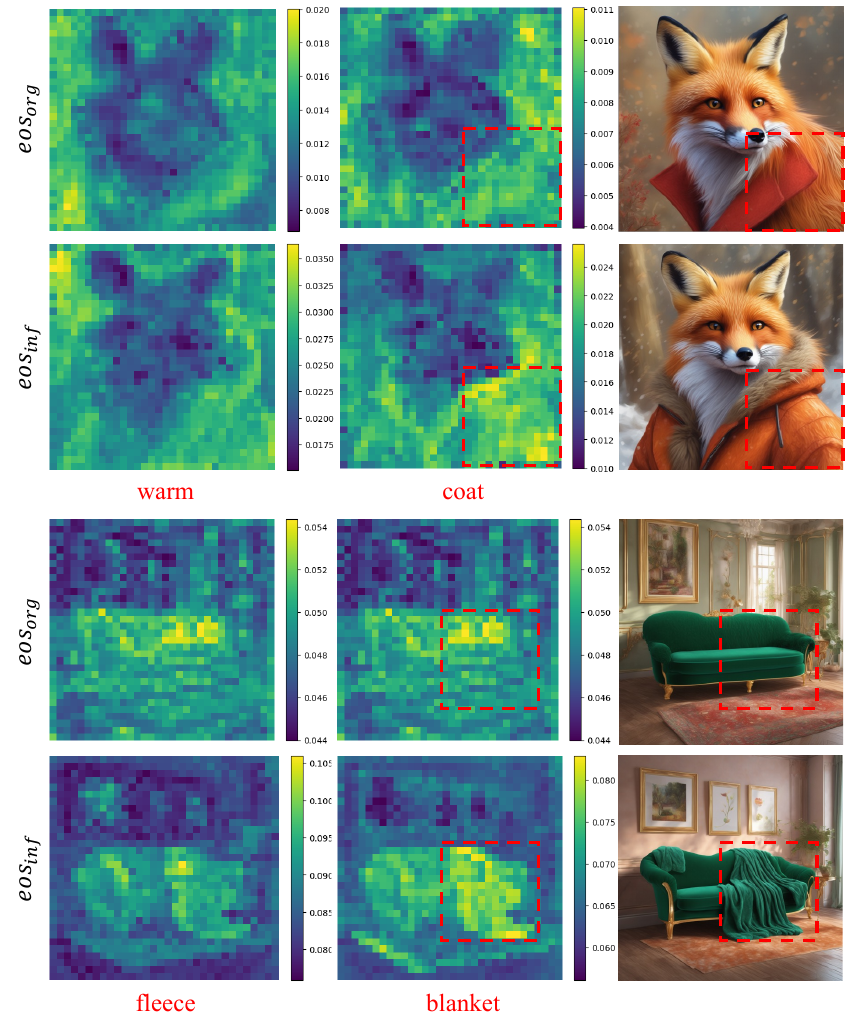}
        \captionof{figure}{
        \textbf{Additional cross-attention map comparison between $eos_{org}$ and $eos_{inf}$.}}
        \label{fig:ca_compare}
    \end{minipage}
\end{figure}

\section{EOS Analysis}
\label{sec:eos_linear}
\noindent\textbf{Empirical Analysis of EOS Linearity.}
Although prior work has shown that CLIP text embeddings often exhibit approximate linear compositionality over semantic factors~\cite{trager2023linear}, it is non-trivial whether the same property holds at the $\langle$EOS$\rangle$ hidden state, which aggregates contextual information through causal attention rather than being a simple sum of word embeddings. Rather than assuming exact linearity, we empirically examine whether the appearance residual $\Delta_{app} = eos_{id} - eos_{sty}$ behaves as a transferable semantic direction. 
To this end, we construct a large prompt set spanning diverse subjects, styles, and scenes, with appearance attributes of varying complexity—from short phrases to long relational descriptions (\cref{fig:eos_linearity}).

\noindent\textit{Parallelism.} We first test whether $\Delta_{app}$ encodes appearance independently of the subject it was extracted from. 
Residuals extracted from the same attribute but different subjects remain highly aligned, whereas residuals from different attributes are clearly separated. 
This indicates that $\Delta_{app}$ captures a subject-agnostic appearance direction rather than subject-specific content.

\noindent\textit{Transferability.} We then add a $\Delta_{app}$ extracted from one subject to the $\langle$EOS$\rangle$ of a different prompt which only contains subject and scene descriptions, and compare it against the embedding of an explicit prompt (subject + appearance + scene) that states the attribute in words (\textit{transfer target}).
The two are highly similar, while both the appearance-free base and a mismatched random attribute are clearly distinguished. 
Notably, this holds even for long, compositional phrases, indicating that EOS-level residuals transfer rich appearance semantics—not merely single words—in a linear manner.

\noindent\textit{Generative effect.} Finally, scaling the injected residual as $eos_{sc} + \lambda \cdot \Delta_{app}$ progressively strengthens the target appearance in the generated image (\cref{fig:eos_linearity}, bottom): as $\lambda$ increases, the attribute (e.g., round glasses) emerges and intensifies, confirming that the residual exerts a controllable, monotonic influence on generation.
Together, these results show that $\Delta_{app}$ functions as a subject-agnostic, transferable, and controllable appearance direction—justifying our EOS-level residual manipulation.

\noindent\textbf{Subspace Construction.}
To compute alignment with appearance and scene subspaces, we first collect all appearance-and scene-related phrases from the Consistory+ benchmark.
To verify whether a meaningful subspace can be constructed, we examine the cumulative explained variance from the PCA of each set. 
As illustrated in \cref{fig:pca_cv}, capturing 90\% of the cumulative variance requires only 81 dimensions for appearance and 221 for scene, out of the total 768 dimensions in the CLIP text embedding space.

    
    

\begin{figure}[!t]
    \centering
    \begin{minipage}[t]{0.35\textwidth}
        \centering
        \includegraphics[width=\linewidth]{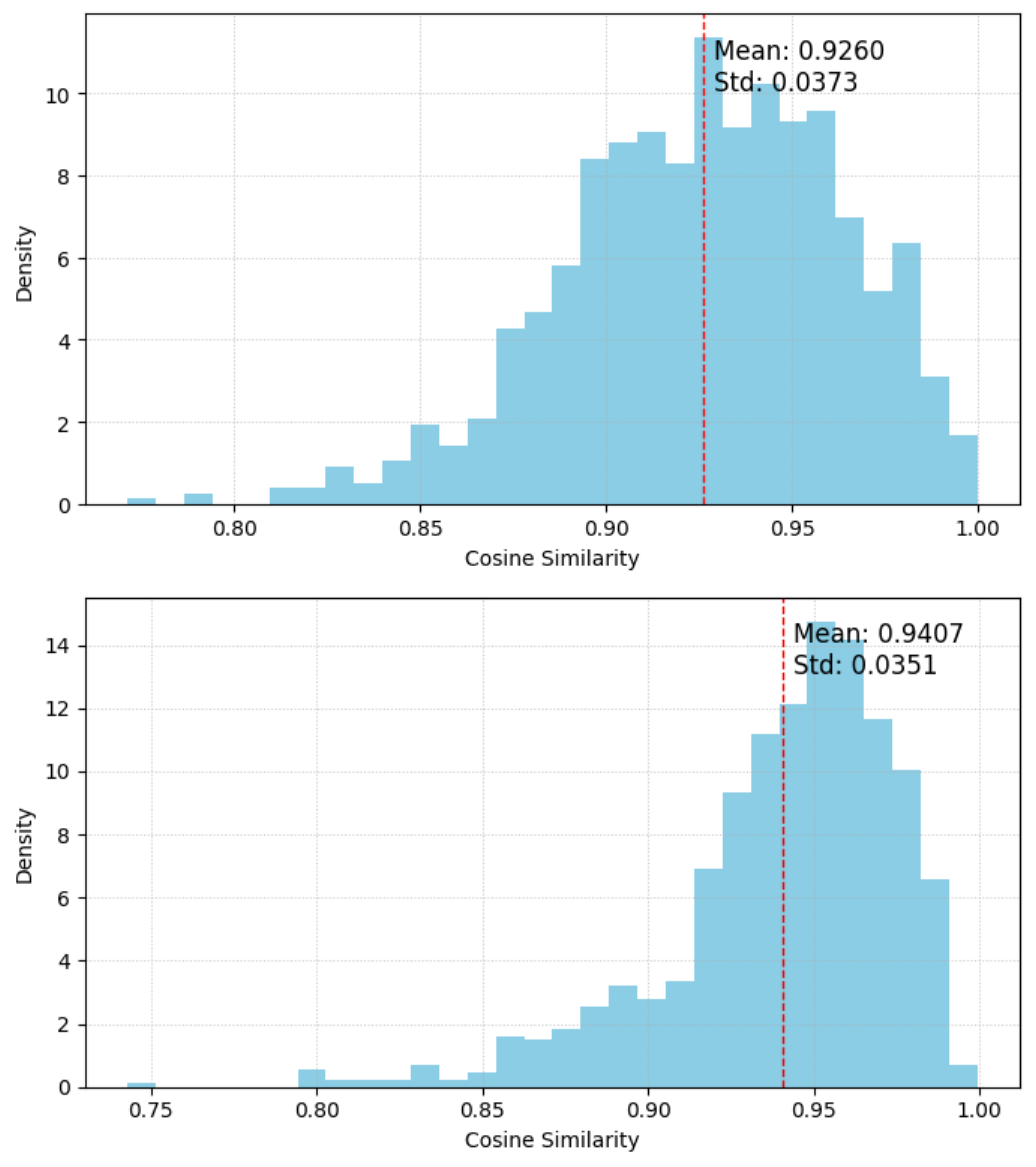}
        \captionof{figure}{\textbf{Cosine similarity distribution between $eos_{org}$ and $eos_{inf}$.}}
        \label{fig:cossim}
    \end{minipage}
    \hfill
    \begin{minipage}[t]{0.6\textwidth}
        \centering
        \includegraphics[width=\linewidth]{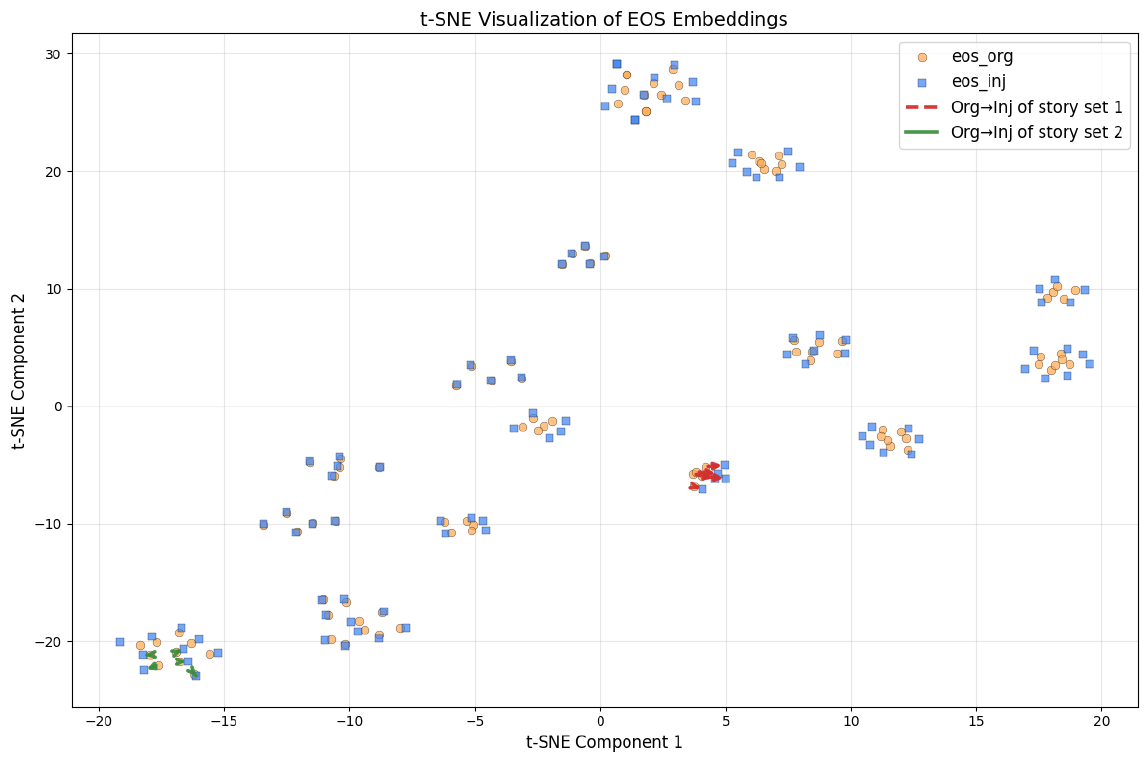}
        \captionof{figure}{\textbf{t-SNE plot of $eos_{org}$ and $eos_{inf}$.}
        Red and green arrows show the position of $eos_{inf}$ from its corresponding $eos_{org}$.}
        \label{fig:tsne}
    \end{minipage}

\end{figure}

\noindent\textbf{Additional Analysis on $eos_{inf}$.}
While $eos_{inf}$ exhibits better alignment with semantic subspaces (\cref{tab:subspace_alignment}), such alignment does not inherently guarantee semantic preservation—unintended deviations from the original meaning may degrade generation quality.
To investigate this, we compute the pairwise cosine similarity between $eos_{org}$ and $eos_{inf}$ across samples (\cref{fig:cossim}), under the hypothesis that significant semantic shifts would result in low similarity.
However, the similarity remains consistently high ($\mu = 0.9260$), suggesting that Cross-Contextual Bridging maintains core semantic content.
We further support this with a t-SNE visualization (\cref{fig:tsne}), where $eos_{inf}$ embeddings consistently cluster near their $eos_{org}$ counterparts.
Together with qualitative examples (\cref{fig:ca_compare}) showing improved attention localization, these results suggest that the infused $\langle$EOS$\rangle$ embedding effectively combines essential semantics while remaining close to the original semantic configuration.



 

\noindent\textbf{Text Encoder Robustness.}
SDXL employs two types of CLIP text encoders: one with a 768 dimensional output and another with 1024 dimensions.
To examine whether the larger encoder also exhibits similar linear behavior, we measure the pairwise cosine similarity between $eos_{org}$ and $eos_{inf}$ using the 1024 dimensional encoder.
As shown in bottom of \cref{fig:cossim}, the similarity remains consistently high ($\mu = 0.9407$), indicating that the Cross-Contextual Bridging mechanism can be used across different text encoder configurations.

\noindent\textbf{Additional Variants of the EOS-Based Control.}
We further analyze variations of EOS-based semantic control in \cref{tab:eos_variation}.
Beyond the main variant presented in the primary experiments (\cref{tab:subspace_alignment}), we systematically examine different EOS construction strategies by varying the base EOS components, injected directions, and mixing weights.
Specifically, $\Delta_{app}$ (\cref{tab:eos_variation}C) corresponds to directly replacing the EOS embedding with the appearance residual, following~\cite{hu2024token}.
The configuration $eos_{sc}^{org}+eos_{app}^{org}+eos_{sty}^{org}$ naively concatenates EOS embeddings obtained from each phrase-level component in the benchmark.
The relatively lower performance of $\Delta_{app}$ can be attributed to the absence of scene information in the resulting embedding.
While another variant $eos_{id} + \Delta_{sc}$ is mathematically similar to our additive formulation, it exhibits slightly reduced alignment. 
We hypothesize that this variant introduces multiple semantic directions originating from overlapping bases, 
whereas our formulation injects a consistent appearance residual $\Delta_{app}$ into multiple scene-grounded $\langle$EOS$\rangle$ anchors $\{eos_{sc}^{n}\}_{n=1}^{N}$.
By shifting several scene embeddings along a consistent semantic direction, our approach maintains a structured integration of narrative context and appearance cues.  
This parallel infusion results in more stable semantic control and ultimately improves text–image alignment compared to subject-anchored alternatives.
Overall, these results indicate that simply replacing or naively combining EOS components does not sufficiently preserve semantic grounding. In contrast, controlled injection of the appearance direction into a scene-aware EOS consistently improves text alignment, highlighting the importance of structured and context-preserving semantic modulation.

\begin{table}[!t]
\centering
\caption{\textbf{$eos_{inf}$ construction ablation.}
We analyze text-alignment (CLIP-T) by varying (A) base EOS 
choice, (B) scaling factor of the injected direction, 
and (C) choice of appearance direction to inject.}
\footnotesize
\setlength{\tabcolsep}{4pt}
\begin{tabular}{l c}
\hline \hline
 & \textbf{CLIP-T} \\ \hline

\multicolumn{2}{l}{\textit{(A) Baselines / naive combinations}} \\ \hline
$eos_{org}$ & 0.8935 \\
$eos_{sc}^{org}+eos_{app}^{org}+eos_{sty}^{org}$ & 0.8625 \\
 $eos_{sc}+eos_{id}$ & 0.8925 \\ \hline

\multicolumn{2}{l}{\textit{(B) Mixing-weight ablation (hyperparameter)}} \\ \hline
 $0.5\,eos_{sc}+0.5\,eos_{id}$ & 0.8864 \\
 $eos_{sc}+0.5\,\Delta_{app}$ & 0.8995 \\
 $eos_{sc}+0.5\,\Delta_{app}^{orth}$ & 0.8992 \\
 $eos_{sc}+1.5\,\Delta_{app}$ & 0.9040 \\
 $eos_{sc}+1.5\,\Delta_{app}^{orth}$ & 0.9038 \\ \hline

\multicolumn{2}{l}{\textit{(C) Direction choice ablation}} \\ \hline
 $\Delta_{app}$ & 0.8726 \\
 $eos_{id}+\Delta_{sc}$ & 0.8865 \\
 $eos_{sc}+\Delta_{app}^{orth}$ & 0.9033 \\ \hline

$eos_{sc}+\Delta_{app}$ (Ours) & \textbf{0.9042} \\
\hline \hline
\end{tabular}
\label{tab:eos_variation}
\end{table}


\begin{table*}[!t]
\centering
\caption{\textbf{Framing phrases in ShotBench}}
\resizebox{1.0\textwidth}{!}{%
\begin{tabular}{ll|ll|ll|ll}
\hline \hline
\textbf{Shot Size} & Wide             & \textbf{Lens Size}      & Long lens            & \textbf{Shot Framing}        & Single            & \textbf{Camera Movement} & Push in           \\
                   & Close up         &                          & Wide                 &                               & Insert            &                         & Pull out          \\
                   & Extreme wide     &                          & Ultra wide (fisheye) &                               & 2 shot            &                         & Boom up           \\
                   & Medium close up  &                          & Medium               &                               & Group shot        &                         & Pan left          \\
                   & Medium wide      & \textbf{Lighting Type}  & Daylight             &                               & Establishing shot &                         & Pan right         \\
                   & Medium           &                          & Artificial light     &                               & Over the shoulder &                         & Tilt down         \\
                   & Extreme close up &                          & Mixed light          &                               & 3 shot            &                         & Tilt up           \\
\textbf{Camera Angle} & Aerial         &                          & Firelight            & \textbf{Lighting Condition}   & Side light        &                         & Boom down         \\
                   & Overhead         &                          & Overcast             &                               & Back light        &                         & Zoom in           \\
                   & Low angle        &                          & Practical light      &                               & High contrast     &                         & Static            \\
                   & High angle       &                          & Sunny                &                               & Silhouette        &                         & Move to right     \\
                   & Dutch angle      &                          & Moonlight            &                               & Edge light        &                         & Move to left      \\
\textbf{Composition} & Center         &                          & Fluorescent          &                               & Underlight        &                         & Zoom out          \\
                   & Balanced         &                          & HMI                  &                               & Top light         &                         & Arc               \\
                   & Symmetrical      &                          & Tungsten             &                               & Hard light        &                         & Camera roll       \\
                   & Right heavy      &                          & LED                  &                               & Soft light        &                         & Dolly zoom        \\
                   & Left heavy       &                          &                      &                               & Low contrast      &                         &                   \\
                   & Short side       &                          &                      &                               &                   &                         &                   \\
\hline \hline
\end{tabular}%
}

\label{tab:framing_phrase}
\end{table*}

\begin{figure*}[!t]
    \centering
    \includegraphics[width=\textwidth]{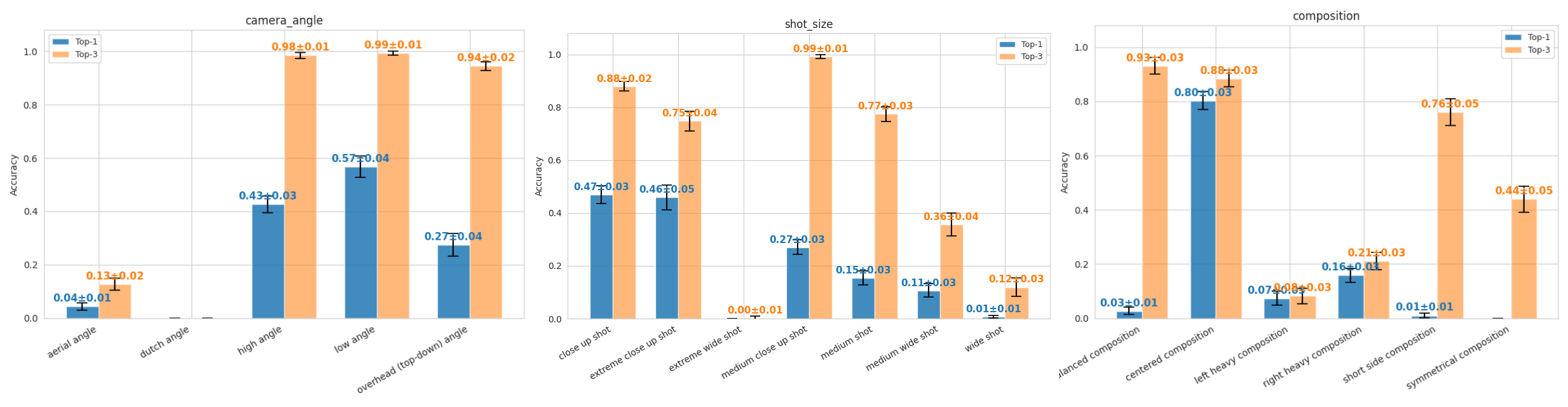}
    \caption{\textbf{Statistical analysis across scene structure categories.}
Each graph presents the mean and standard deviation of top-1 and top-3 accuracies for each option within a category. Results are averaged over 10 random seeds.}
    \label{fig:graph_shot}
\end{figure*}

\section{Consistory+\_+ \& Framing Evaluation Details}



The ShotBench~\cite{liu2025shotbench} includes a wide range of framing phrases across eight categories (\cref{tab:framing_phrase}). 
To ensure a focused and rigorous evaluation of scene-specific structural variance, we intentionally restricted the scope to three primary categories, excluding the remaining five based on the following methodological criteria: (1) Prompt Redundancy: Shot Framing descriptors (e.g., ``single'', ``2 shot'') primarily specify the number of characters, which is already inherently defined by the subject phrases in the storyboard prompts; (2) Redundancy Mitigation: Lens Size phrases (e.g., ``wide lens'') are largely redundant as they are semantically encompassed by the Shot Size category (``wide shot''); (3) Domain Incompatibility: Camera Movement is inherently applicable to temporal video sequences and is thus irrelevant to static image generation; and (4) Task Relevance: Lighting-related categories fall outside the scope of our work, as they primarily focus on stylistic or artistic variations rather than the structural and compositional elements that define scene-specific variance.

\noindent\textbf{Framing Phrase Filtering.} Although SDXL generally produces textually well-aligned images, it sometimes struggles to accurately reflect specific framing phrases.
Before incorporating these phrases into our benchmark extension, we first evaluate how well they are captured by the base SDXL model using the ShotBench evaluation mechanism.
To quantitatively assess phrase reflection, we construct approximately 90 diverse scene prompts and append each framing phrase at the end (e.g., ``A photo of a woman baking cookies, close-up shot'').
We generate images for 10 random seeds per phrase and compute the top-1 and top-3 accuracy, along with standard deviations, as shown in \cref{fig:graph_shot}.
For example, in the ``shot size'' category with 7 phrase options, this results in a total of 6,300 images (90 scenes × 7 phrases × 10 seeds).
We filter out framing phrases whose top-1 accuracy is zero (e.g., ``extreme wide shot''), as these are not reliably reflected in the generated outputs from SDXL.

\noindent\textbf{Consistory+\_+}
With the remaining framing phrases, we extend the original Consistory+ benchmark using GPT-4o.
Since these phrases are visually meaningful primarily for animate subjects, we first guide GPT-4o to determine whether the subject in each prompt is animate.
Then, the model is instructed to append the most appropriate framing phrase for the given scene, resulting in Consistory+\_+.

\noindent\textbf{Framing Accuracy Evaluation.}
Following the evaluation protocol in~\cite{liu2025shotbench}, we provide the generated image along with the corresponding framing category (e.g., shot size) and prompt the model with a multiple-choice question.
For instance, for shot size evaluation, we prompt the model with a multiple-choice question such as 
``What is the shot size of this shot?'' and provide predefined options (e.g., wide shot, medium shot, close-up shot). 
The predicted category is compared against the framing phrase specified in the text prompt, and accuracy is reported.

\section{Additional Experiments}

\subsection{Ablation Studies}
\begin{figure}[!t]
\centering
\includegraphics[width=\linewidth]{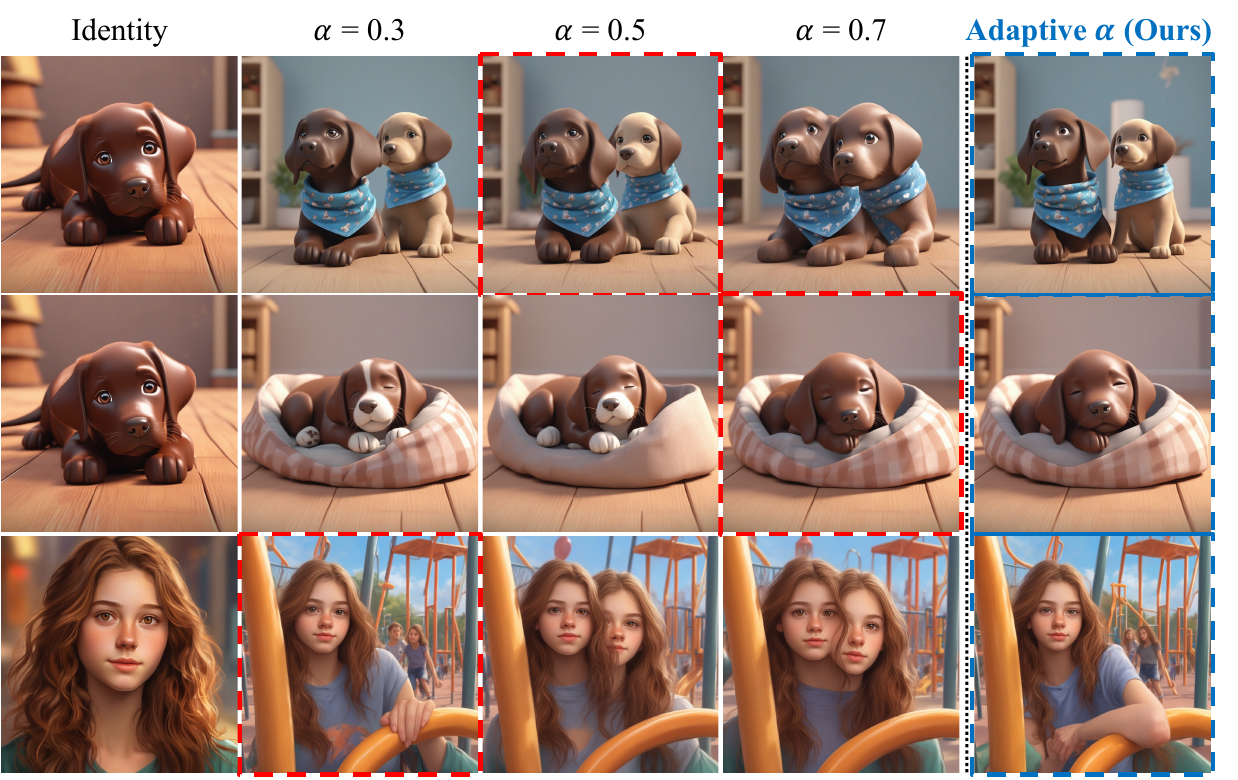}
\caption{\textbf{Effectiveness of adaptive thresholding.} 
While different scenes require distinct $\alpha$ values (red), our adaptive threshold ($\mu + \sigma$) provides reliable results by dynamically adjusting to each scene (blue).}
\label{fig:alpha}
\end{figure}

\begin{figure}[!t]
\centering

\begin{minipage}[t]{0.52\textwidth}
\vspace{0pt}
\centering
\includegraphics[width=\linewidth]{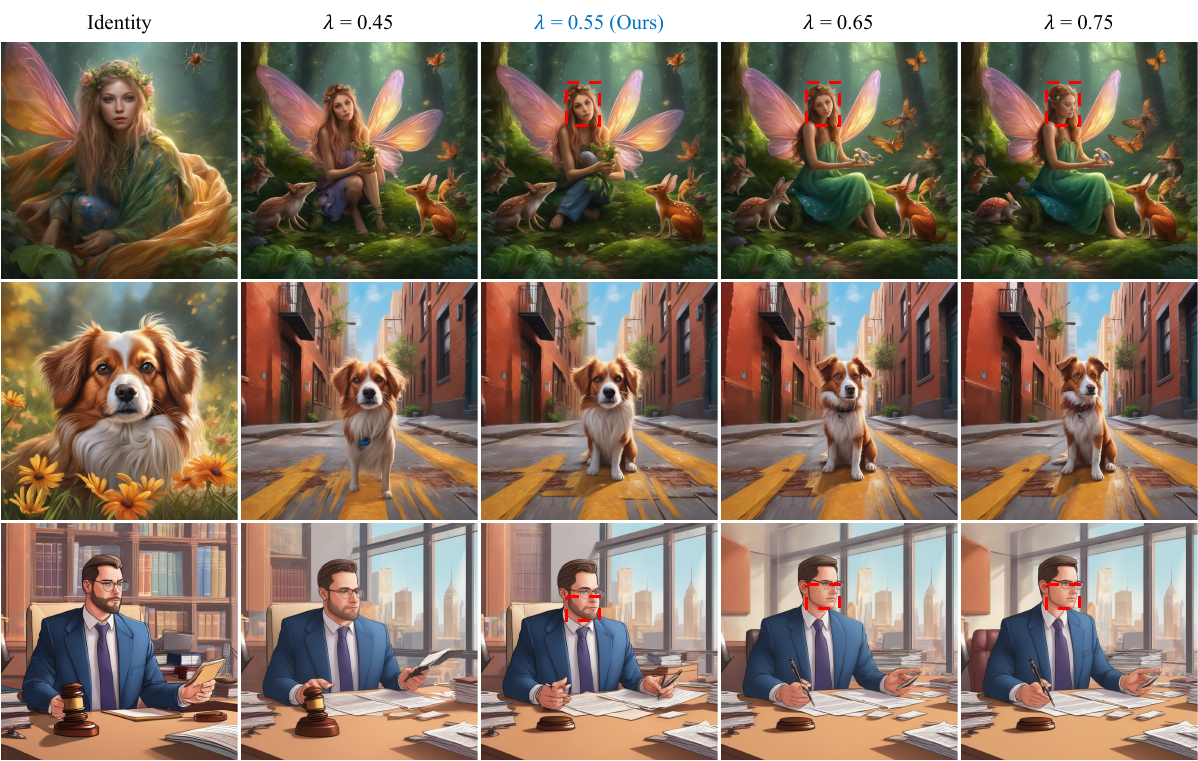}
\captionof{figure}{\textbf{Visualization of $\lambda$ sensitivity.} As $\lambda$ increases, the degree of subject identity preservation decreases.}
\label{fig:hyper_vis}
\end{minipage}
\hfill
\begin{minipage}[t]{0.46\textwidth}
\vspace{0pt}
\centering
\captionof{table}{\textbf{Ablation of $\alpha$ and blending scale $\lambda$}}
\label{tab:hyperparameter}

\setlength{\tabcolsep}{3pt}
\renewcommand{\arraystretch}{0.95}
\scriptsize

\resizebox{\linewidth}{!}{%
\begin{tabular}{c|cc|c|cc}
\hline \hline
$\alpha$    & CLIP-T & CLIP-I & $\lambda$ & CLIP-T & CLIP-I \\ \hline
Baseline & 0.8971 & 0.8430 & - & - & - \\ \hline
0.1      & 0.9033 & 0.8841 & 0.45      & 0.9027       &   0.8851     \\
0.2      & 0.9036 & 0.8836 & 0.5       & 0.9033       &    0.8843    \\
0.3      & 0.9039 & 0.8848 & 0.55      & 0.9042 & 0.8850  \\
0.4      & 0.9040  & 0.8852 & 0.6       & 0.9044 & 0.8837 \\
0.5      & 0.9033 & 0.8860  & 0.65      & 0.9034 & 0.8797 \\
0.6      & 0.9026 & 0.8864 & 0.7       & 0.9035 & 0.8798 \\
0.7      & 0.9022 & 0.8871 & 0.75      & 0.9038 & 0.8787 \\
$\mu + \sigma$ & 0.9042 & 0.8850  & 0.8       & 0.9044 & 0.8769 \\ \hline \hline
\end{tabular}%
}
\end{minipage}

\vspace{-2mm}
\end{figure}

\noindent\textbf{Top-$\alpha$ and Blending Scale $\lambda$.}
\cref{fig:hyper_vis} shows that as $\lambda$ decreases (thereby injecting more subject-relevant features), the generated results better capture the identity characteristics (e.g., beard of a man).
In \cref{tab:hyperparameter}, we report the performance variations with respect to different hyperparameter settings.
We set $\alpha = \mu + \sigma$ and $\lambda = 0.55$ based on the best trade-off observed between the CLIP-T and CLIP-I score.
Furthermore, \cref{fig:alpha} illustrates the effect of similarity filtering (\cref{sec:identityanchoring}). 
While a fixed proportion $\alpha$ can preserve identity, varying similarity distributions across scenes may cause noisy or discarded correspondences, leading to artifacts such as duplicated features. 
In contrast, our adaptive threshold derives scene-specific criteria from the similarity distribution, enabling more stable identity integration without such artifacts.



\begin{figure}[!t]
\centering

\begin{minipage}[t]{0.48\textwidth}
\vspace{-30mm}
    \centering
       \captionof{table}{\textbf{EOS substitution range.} 
    We compare the result when change all the remaining EOS embeddings in text embeddings.}
\resizebox{0.95\linewidth}{!}{%
\begin{tabular}{l|cc}
\hline \hline
           & All  &  First (Ours) \\ \hline
CLIP-I     & 0.8741         & \textbf{0.8850}         \\
Dreamsim   & 0.2185         & \textbf{0.2005}         \\
Aesthetic  & 6.1111         & \textbf{6.1764}         \\ \hline \hline
\end{tabular}%
}
    \label{tab:change-range}

\end{minipage}
\hfill
\begin{minipage}[t]{0.5\textwidth}
    \centering
    
    \includegraphics[width=\linewidth]{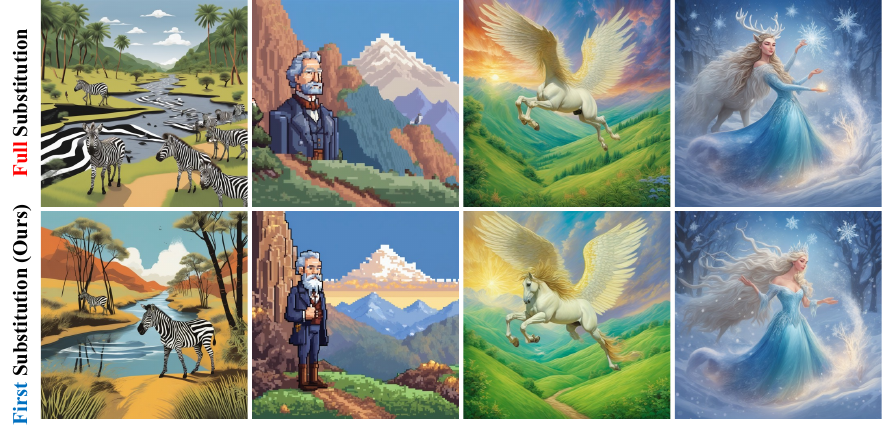}
    
    \captionof{figure}{\textbf{Image comparison of EOS substitution scope.} 
    Full substitution often leads to misaligned generations, resulting in unnatural scene rendering.}
    \label{fig:change-range}

\end{minipage}

\end{figure}

\noindent\textbf{EOS Change Range.}
Since text embeddings include multiple $\langle$EOS$\rangle$ tokens after the sentence ends, we investigate the effect of the substitution range in \cref{tab:change-range}. 
The results show that replacing all $\langle$EOS$\rangle$ tokens often degrades image quality and leads to unnatural scene rendering (\cref{fig:change-range}), likely due to redundant perturbations introduced when modifying padded EOS positions.



\begin{figure}[!t]
    \centering
    \begin{minipage}{0.47\textwidth}
        \centering
        \includegraphics[width=\linewidth]{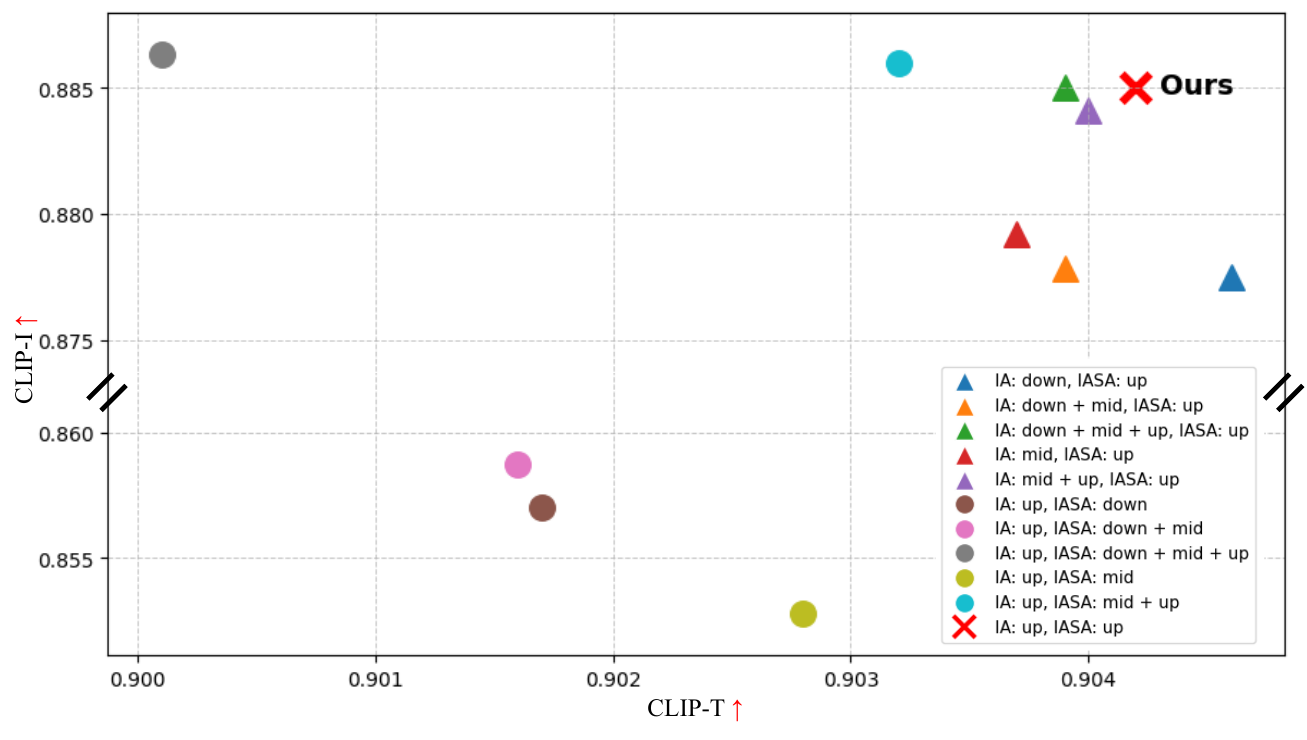}
        \caption{\textbf{Score difference across injection layers.} 
        We visualize the score changes when applying IA and IASA at different layers.
        Higher values toward the top-right indicate better overall performance.}
        \label{fig:layer-change}
    \end{minipage}
    \hfill
    \begin{minipage}{0.47\textwidth}
        \centering
        \includegraphics[width=\linewidth]{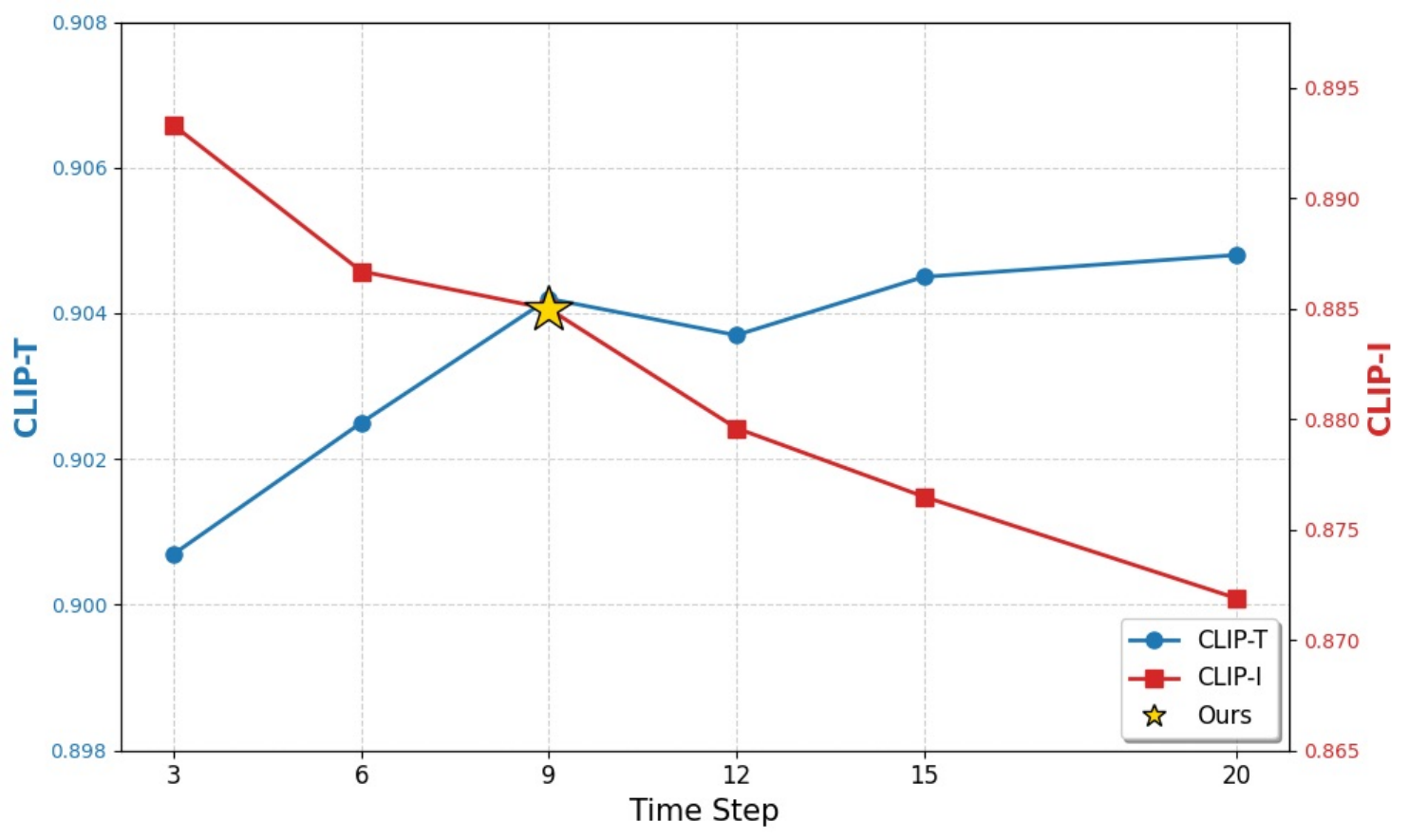}
        \caption{\textbf{Performance trade-off with switching step.} 
        Switching at $t=9$ effectively balances CLIP-T and CLIP-I scores.}
        \label{fig:timestep}
    \end{minipage}
\end{figure}

\noindent\textbf{Layer Application Variation.}
We apply Identity Anchoring (IA) and Identity-Aware Self-Attention (IASA) at different layers of the unet decoder in the diffusion model.
While prior works typically transfer identity-related features within the decoder (up-block) layers, we further investigate the impact of layer selection.
As shown in \cref{fig:layer-change}, we vary the application layers and plot the resulting CLIP-T and CLIP-I scores.
Interestingly, when IA and IASA are applied to configurations that include the up-block layers, the CLIP-I score notably improves (surpassing 0.880).
These results empirically highlight the importance of the decoder layers (up-blocks) in effectively conveying identity features and enhancing the image quality during generation.

\noindent\textbf{Switching Timestep in AIP.}
Prior research indicates that the global layout of generated images is primarily established during the initial denoising stages. 
Based on this observation, we evaluate the performance trade-offs by varying the switching timestep between Identity Anchoring (IA) and Identity-Aware Self-Attention (IASA).
As illustrated in \cref{fig:timestep}, transitioning at $t=9$ effectively balances prompt alignment (CLIP-T) and identity consistency (CLIP-I), ensuring both structural flexibility and identity preservation.



\subsection{Extended Experiments}
\begin{figure}[!t]
\centering

\begin{minipage}[t]{0.48\textwidth}
\vspace{0pt}
\centering
\includegraphics[width=\linewidth]{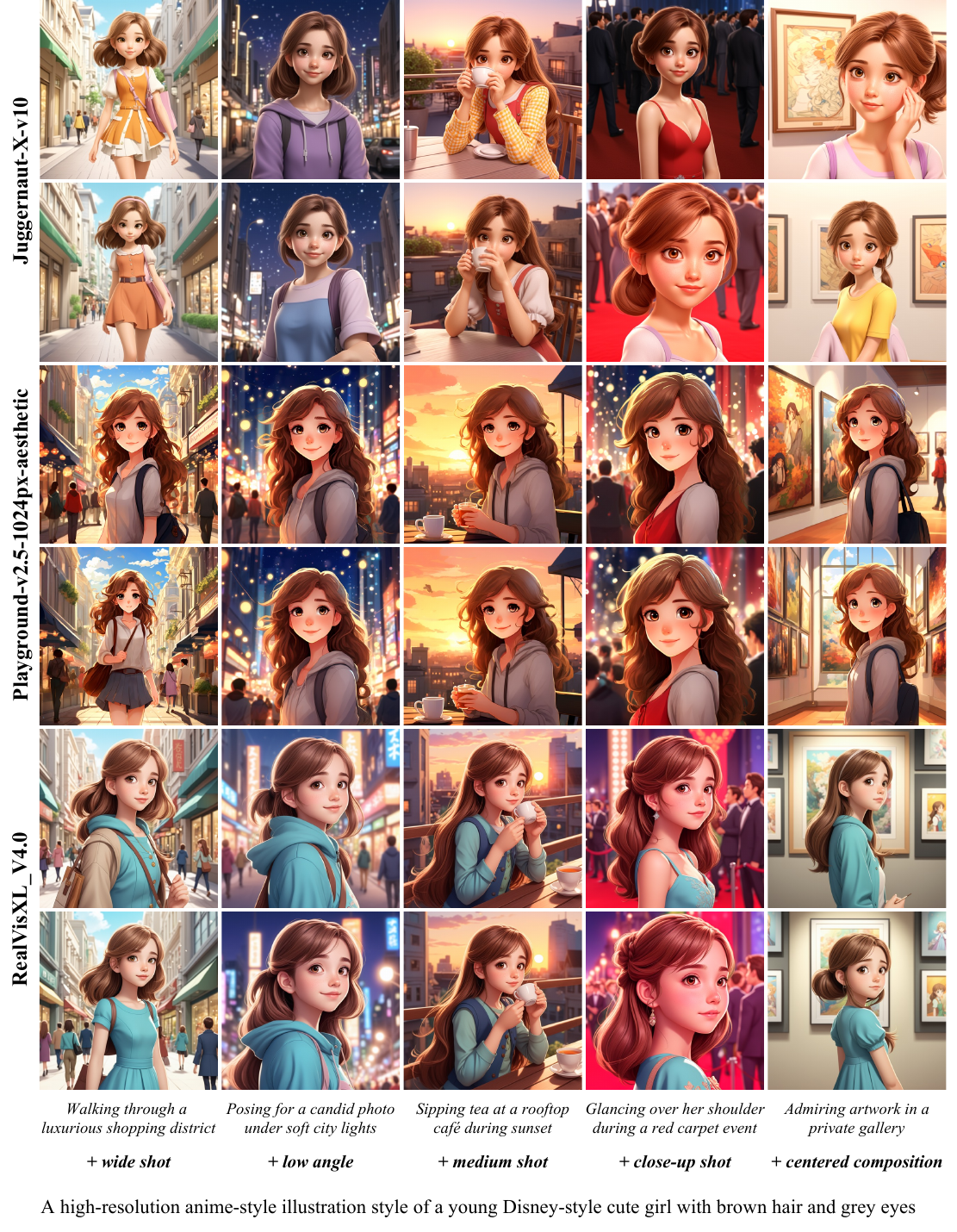}
\captionof{figure}{\textbf{Results on different model weights.} 
For each row in every model, framing phrases are added to the end of prompts.}
\label{fig:model_variation}
\end{minipage}
\hfill
\begin{minipage}[t]{0.48\textwidth}
\vspace{0pt}
\centering

\includegraphics[width=\linewidth]{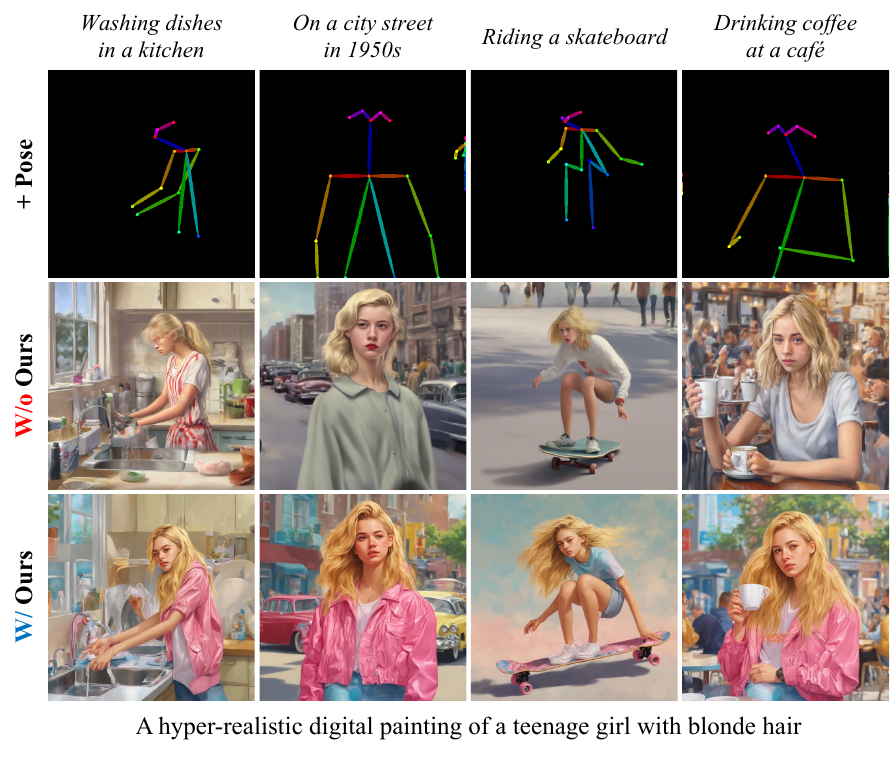}
\captionof{figure}{\textbf{Image results with ControlNet.} 
Our model can be integrated with various conditioning models, such as ControlNet.}
\label{fig:controlnet}

\vspace{3mm}

\captionof{table}{Score on Consistory+ on SD3.}
\resizebox{\linewidth}{!}{%
\begin{tabular}{l|ccccc}
\hline
 & CLIP-T & CLIP-I  & Dreamsim \\ \hline
SD3        & 0.8786 & 0.8517  & 0.3076 \\
Ours (SD3) & \textbf{0.8978} & \textbf{0.8728}  & \textbf{0.2571} \\ \hline
\end{tabular}
}
\label{tab:sd3_exp}

\end{minipage}

\end{figure}

\noindent\textbf{Model Variation.}
To evaluate the generalizability of our method across different model variants based on the SDXL architecture,
we apply EOS-Story to several pretrained weights: Juggernaut-X-v10, Playground-v2.5, and RealVisXL.
We generate images both with and without framing phrases in \cref{fig:model_variation}.
Although the influence of framing phrases may appear subtle in some cases—particularly when the original scene already shares similar structural layouts—the generated images consistently reflect the intended semantics.

\noindent\textbf{ControlNet Adaptation.}
Our model can also be adapted to work with various pretrained conditioning models such as ControlNet.
As shown in \cref{fig:controlnet}, integrating our method with ControlNet enables the generation of subject-coherent image sequences with diverse poses.

\noindent\textbf{Free-form Prompt.}
Following prior works, our model adopts the structured prompt format used in the Consistory+ benchmark. However, in practical
\begin{wrapfigure}[11]{r}{0.54\textwidth}
\vspace{-4mm}
\centering
\includegraphics[width=\linewidth]{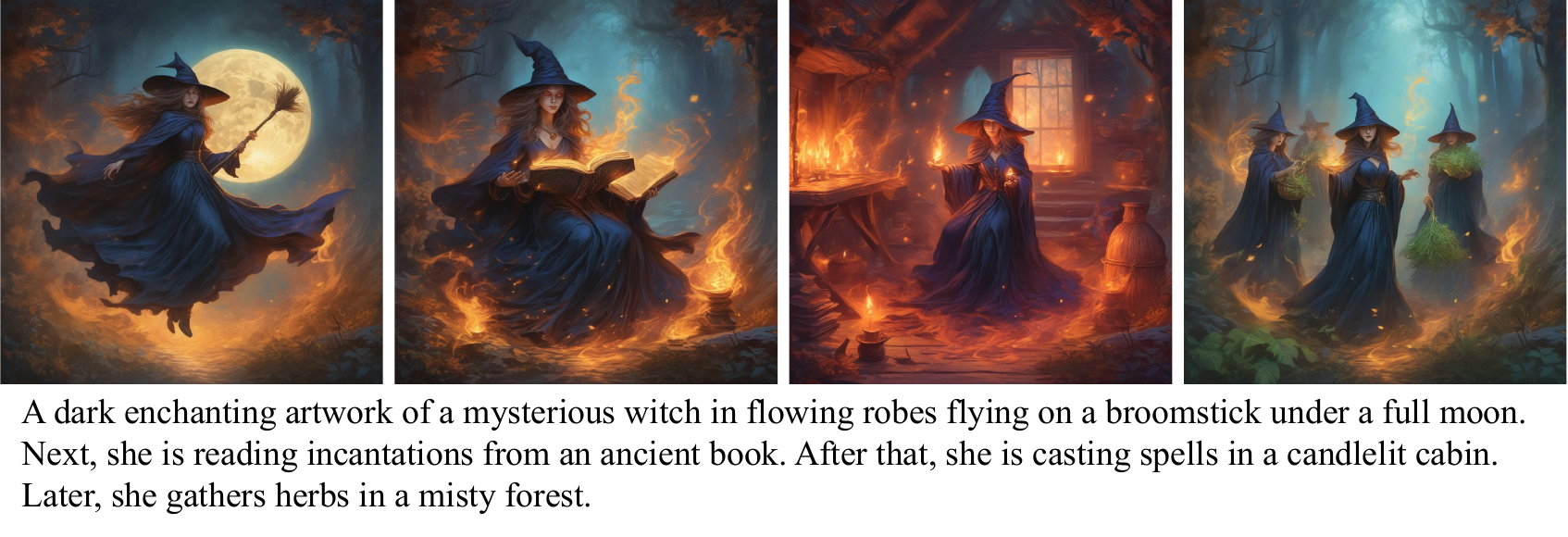}
\caption{\textbf{Free-form prompt generation results.}
Our model can handle free-form user prompts by parsing them into the intended EOS-Story format with an LLM.}
\label{fig:freeform}
\vspace{-4mm}
\end{wrapfigure}
scenarios, users may provide free-form natural language prompts instead of carefully formatted inputs. To bridge this gap, we additionally employ GPT-4o as a prompt parser that converts free-form user instructions into the intended prompt format for EOS-Story, thereby enabling our method to support more flexible user inputs, as shown in \cref{fig:freeform} and \cref{tab:freeform_score}.

\begin{table}[!h]
\vspace{-4mm}
\centering
\caption{\textbf{Quantitative results on free-form story generation.}}
\label{tab:freeform_score}
\resizebox{\linewidth}{!}{
\begin{tabular}{lccccc}
\toprule
\textbf{Method} 
& \textbf{CLIP-T$\uparrow$} 
& \textbf{CLIP-I$\uparrow$} 
& \textbf{DreamSim$\downarrow$} 
& \textbf{Aesthetic$\uparrow$} 
& \textbf{ImageReward$\uparrow$} \\
\midrule
SDXL (Base) & 0.8971 & 0.8430 & 0.3124 & 6.1227 & 0.8289 \\
EOS-Story (Ours) & \textbf{0.9042} & \textbf{0.8850} & \underline{0.2005} & \underline{6.1764} & \underline{1.1680} \\
EOS-Story (Free-form) & \underline{0.9040} & \underline{0.8848} & \textbf{0.2001} & \textbf{6.1867} & \textbf{1.1712} \\
\bottomrule
\end{tabular}
}
\end{table}

\begin{figure}[!h]
    \centering
    \includegraphics[width=0.95 \linewidth]{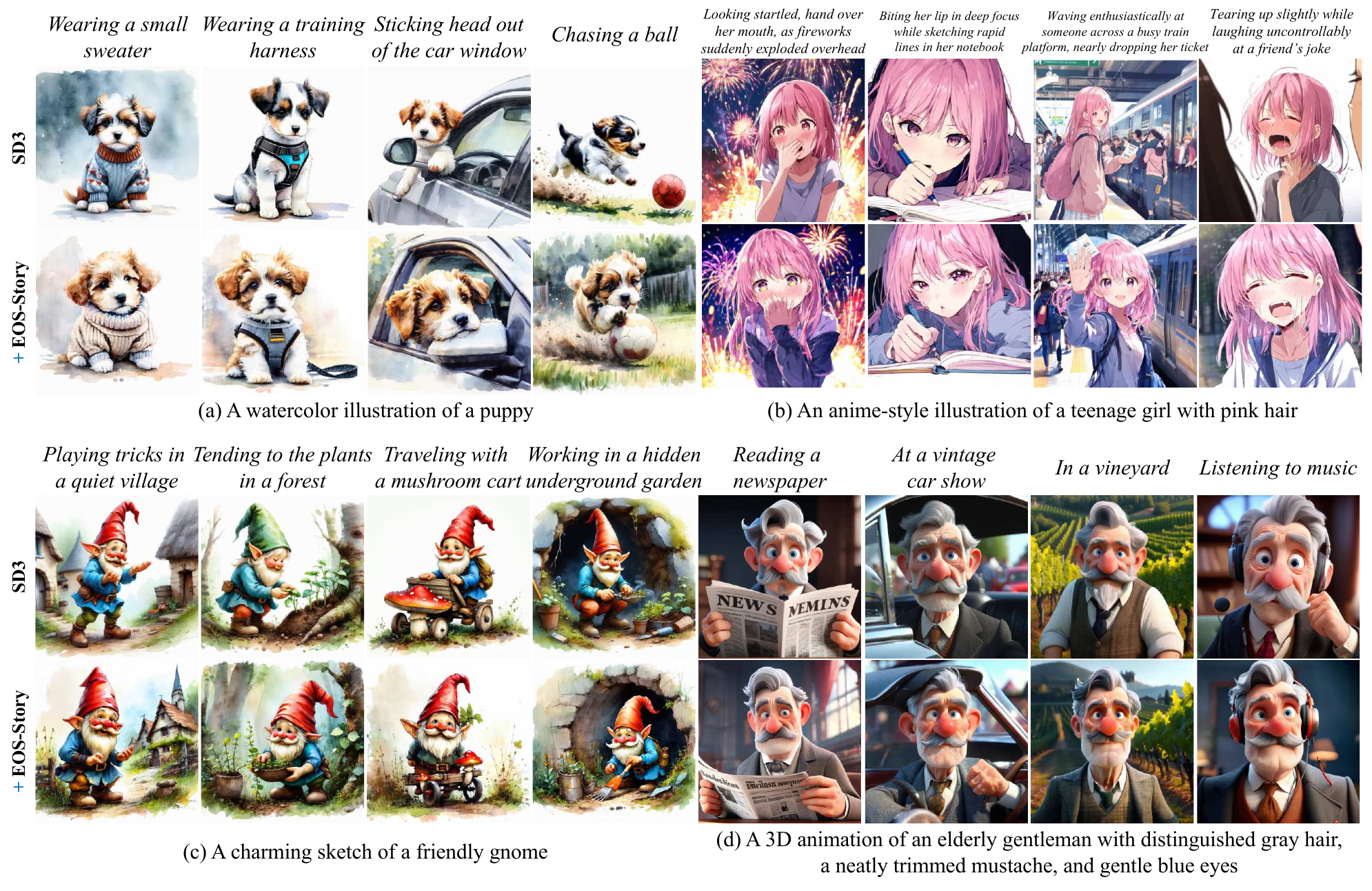}
    \caption{\textbf{EOS-Story results with SD3~\cite{esser2024scaling}.}}
    \label{fig:sd3_supp}
\end{figure}

\noindent\textbf{SD3 Applicability.}
We use the Stable Diffusion 3.5 Large weights with a CFG scale of 4.5 and 28 sampling steps.
In SDXL, identity information is incorporated via extended attention, where the subject key and value embeddings are concatenated with those of the scene. However, in DiT-based models, fixed positional encodings and sequence-length constraints make such direct concatenation non-trivial~\cite{ding2024longrope, lu2024fit}.
To address this architectural difference, we compute attention along two parallel paths: (i) the original scene query–key–value path and (ii) a subject-referenced path using the scene query with subject key and value embeddings. The resulting attention outputs are then combined, enabling implicit identity referencing without modifying the positional encoding structure.
Despite the architectural adaptation, the core semantic control mechanism remains intact. As shown in \cref{fig:sd3_supp} and \cref{tab:sd3_exp}, our method consistently improves alignment, demonstrating its applicability to DiT-based diffusion models.

\noindent\textbf{FLUX Applicability.}
Although SD3 also adopts MM-DiT blocks, we further extend our method to FLUX, as many recent models are built upon it. Since FLUX relies solely on a T5 encoder—unlike the CLIP encoder used in the SD series—the ⟨EOS⟩ token acts merely as an end-of-sentence marker with negligible influence on generation. We therefore apply only the AIP module to FLUX, which is model-agnostic by design.
\begin{table}[!t]
\centering
{\footnotesize
\setlength{\tabcolsep}{4pt}
\renewcommand{\arraystretch}{1.15}
\begin{tabular}{l|ccccc}
\hline
                & CLIP-T$\uparrow$ & CLIP-I$\uparrow$ & DreamSim$\downarrow$ & ImageReward$\uparrow$ & Aesthetic$\uparrow$ \\ \hline
FLUX (baseline) & 0.8696 & 0.8555 & 0.2445 & 1.2988 & 6.0466 \\ \hline
\multicolumn{6}{c}{\textit{\textcolor{red}{Large Pretrained}}} \\ \hline
\rowcolor{gray!15} FLUX-Kontext    & 0.8544 & \textbf{0.9032} & \textbf{0.1602} & 1.2246 & 5.9489 \\ \hline
\multicolumn{6}{c}{\textit{\textcolor{green!60!black}{Training-Free}}} \\ \hline
CharaConsist    & \underline{0.8697} & 0.8636 & 0.2358 & \textbf{1.2774} & \underline{6.0572} \\
EOS-Story (FLUX) & \textbf{0.8722} & \underline{0.8712} & \underline{0.2052} & \underline{1.2673} & \textbf{6.0943} \\ \hline
\end{tabular}
}
\caption{\textbf{Quantitative comparison under the FLUX backbone.} \textbf{Bold} and \underline{underline} denote the best and second-best.}
\label{tab:flux_exp}
\end{table}

\begin{figure}[!t]
    \centering
    \includegraphics[width=0.55\textwidth]{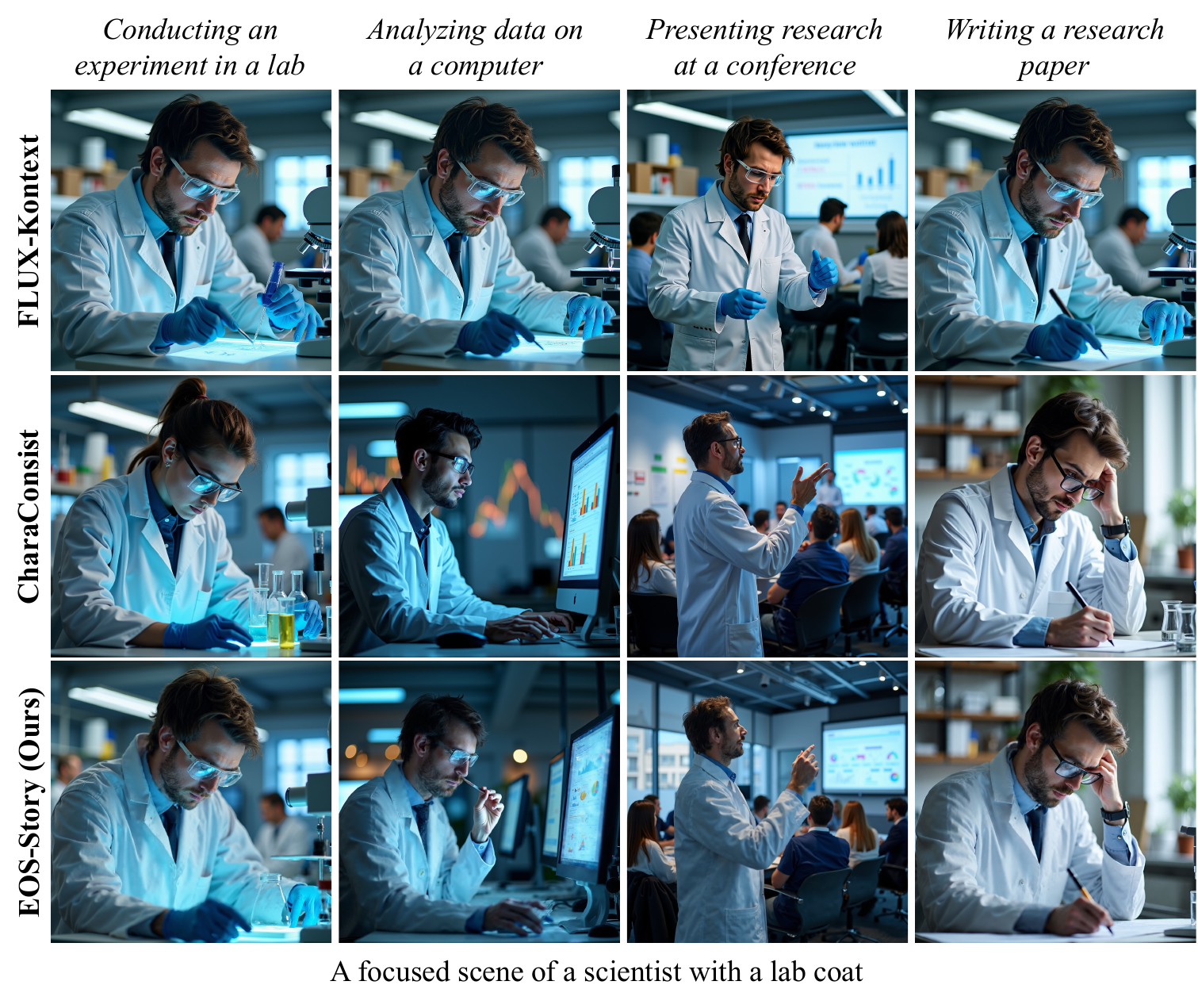}
    \caption{\textbf{EOS-Story on FLUX.} Although the large pre-trained model (Flux-Kontext) better preserves the subject identity, it still struggles to reflect the given text (e.g., computer) and produces rigidly similar pose.}
    \label{fig:flux}
\end{figure}
Recent large pretrained image editing model FLUX-Kontext exhibits strong identity-preservation capabilities and can be directly applied to visual storytelling. 
However, we observe that it remains anchored to the reference, producing copy-and-paste-style synthesis that fails to follow scene-specific prompts (\cref{fig:flux}, Tab.~\ref{tab:flux_exp}).
As shown in Tab.~\ref{tab:flux_exp}, while it achieves strong identity scores (e.g., CLIP-I, DreamSim), its prompt alignment (CLIP-T) and overall quality remain limited, indicating that identity preservation alone is insufficient for storytelling—scene-specific adaptation is essential.
We further compare with CharaConsist~\cite{Wang_2025_ICCV}, a recent FLUX-based storytelling method. Since its evaluation benchmark is not publicly available, we compare on Consistory+ for a fair setting. EOS-Story achieves competitive or superior performance across both identity and prompt-alignment metrics, demonstrating that our scene-adaptive design transfers effectively to FLUX.


\begin{figure}[!h]
    \centering
    \includegraphics[width=0.55\textwidth]{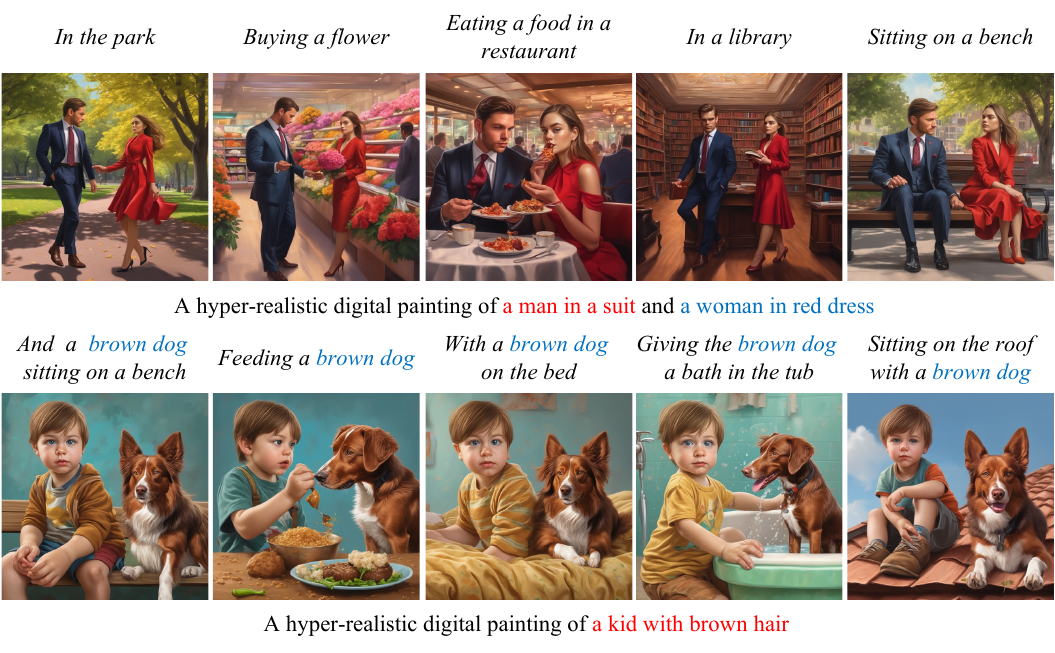}
    \caption{\textbf{Image results on multi-characters.}}
    \label{fig:multi}
\end{figure}

\noindent\textbf{Multi-Character Generation.}
By defining multiple characters, our model can be applied to generate images that preserve each subject’s identity.
As shown in \cref{fig:multi}, it captures both the intended prompt and subject identities, resulting in coherent and visually consistent story sequences.

\begin{figure}[!t]
\centering
\begin{minipage}[t]{0.50\linewidth}
\vspace{0pt}
\centering
\includegraphics[width=0.65\linewidth]{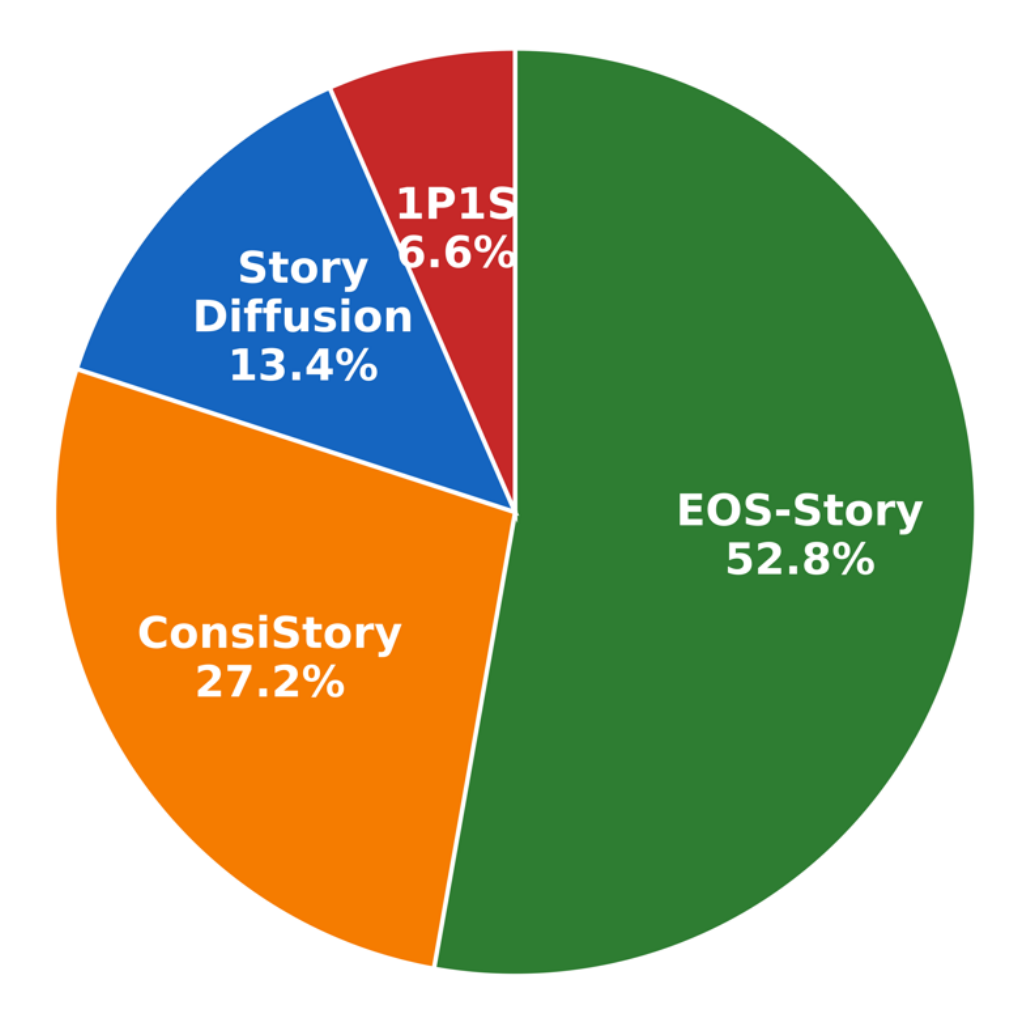}
\captionof{figure}{\textbf{User study results.} EOS-Story is most preferred (52.8\%) for balancing identity, prompt alignment, and scene adaptability across 20 participants.}
\label{fig:user_result}

\vspace{3mm}

\small
\setlength{\tabcolsep}{3pt}
\begin{tabular}{lcc}
\toprule
& GPU & Latency \\
& (MB) & (s) \\
\midrule
StoryDiffusion & 46287 & 98.774 \\
Consistory & 43628 & 144.814 \\
1P1S & 17073 & 118.739 \\
EOS-Story & 47777 & 118.570 \\
\bottomrule
\end{tabular}
\vspace{-1mm}
\captionof{table}{Cost of generating 5 stories.}
\label{tab:efficiency}
\end{minipage}
\hfill
\begin{minipage}[t]{0.48\linewidth}
\vspace{0pt}
\centering
\includegraphics[width=\linewidth]{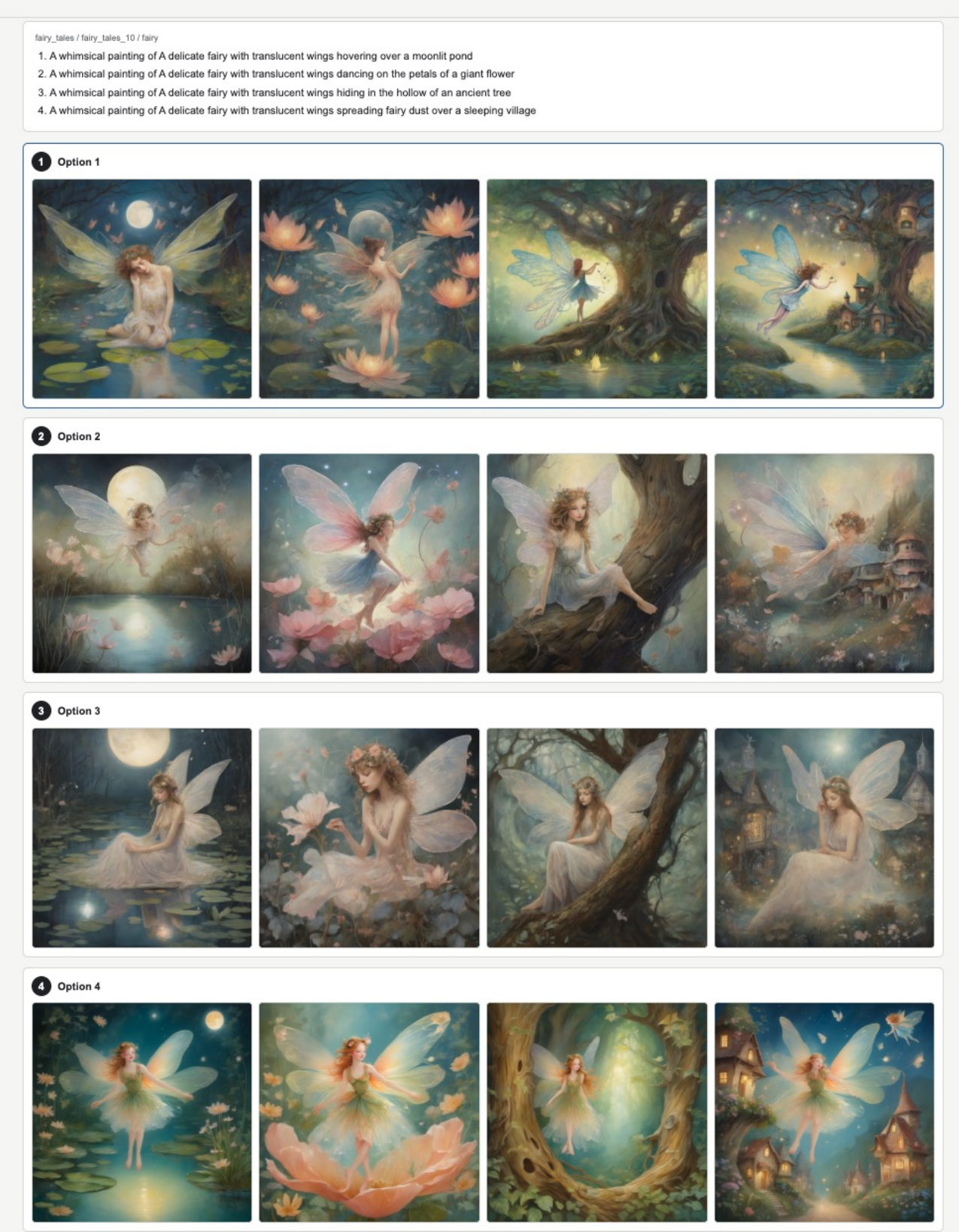}
\captionof{figure}{\textbf{User study example.} Participants were shown four options (one per method) generated from the same storyboard, and asked to select the result that best balances identity, prompt alignment, and scene adaptability.}
\label{fig:user_example}
\end{minipage}
\end{figure}

\noindent\textbf{User Study \& Computational Cost.}
We conducted a user study with 20 participants based on 40 example storyboards. For each storyboard, participants were asked to select the result that best balances identity consistency, prompt alignment, and scene adaptability among the four methods (\cref{fig:user_example}). As shown in \cref{fig:user_result}, EOS-Story was selected as the best in 52.8\% of cases, outperforming the baselines.
We report GPU memory usage and latency for generating 5 stories in \cref{tab:efficiency}. Note that StoryDiffusion is measured at 768$\times$768 resolution, and 1P1S processes images individually rather than in batches, resulting in lower GPU usage. While EOS-Story uses slightly more memory than the baselines, its latency remains comparable, indicating that the additional memory cost does not translate into a noticeable runtime overhead.

\subsection{Additional Qualitative Results}
\begin{figure*}[!t]
    \centering
    \includegraphics[width=1.0\textwidth]{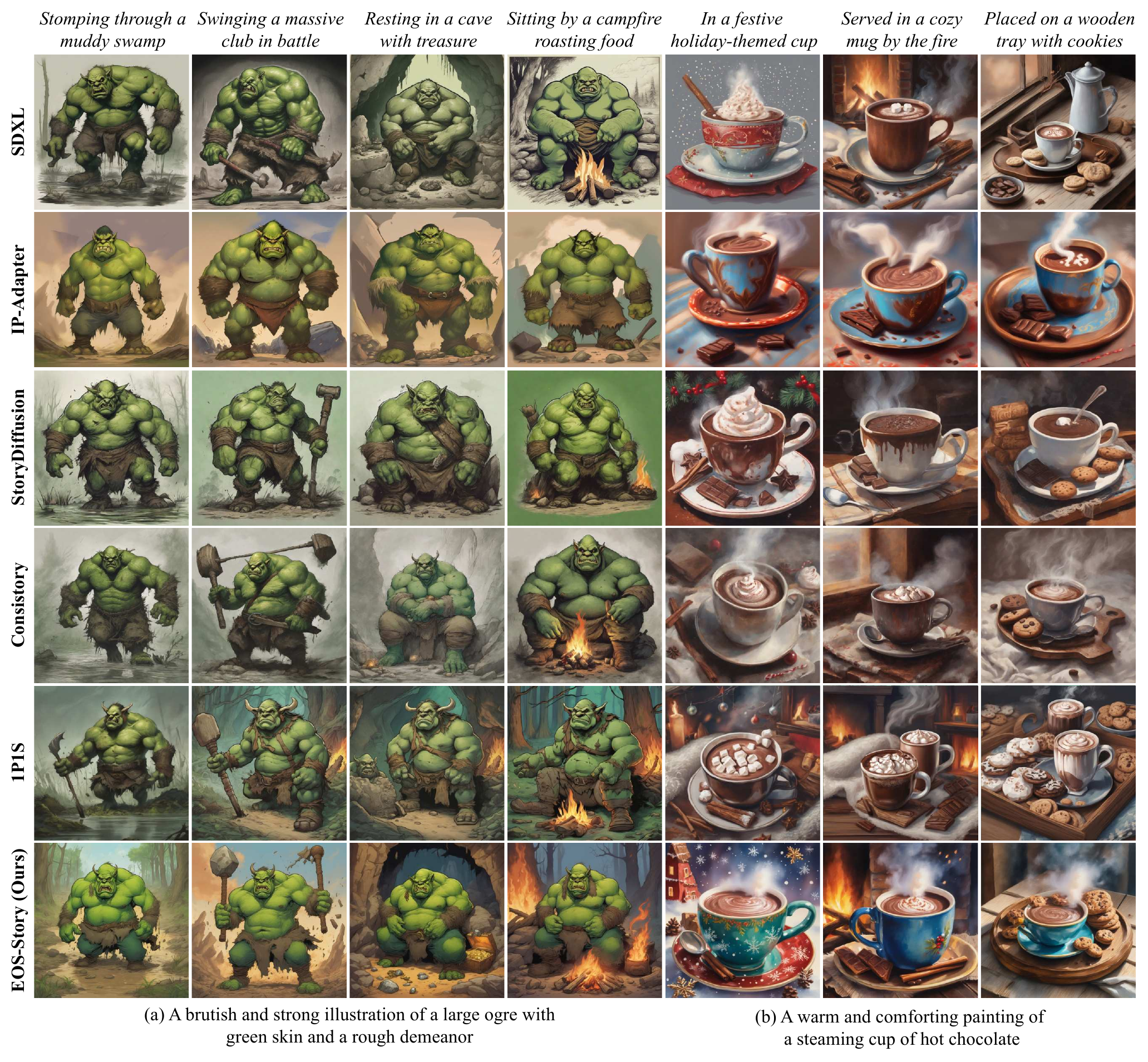}
    \caption{\textbf{Additional qualitative comparison.}}
    \label{fig:add_compare2}
\end{figure*}

\begin{figure*}[!t]
    \centering
    \includegraphics[width=1.0\textwidth]{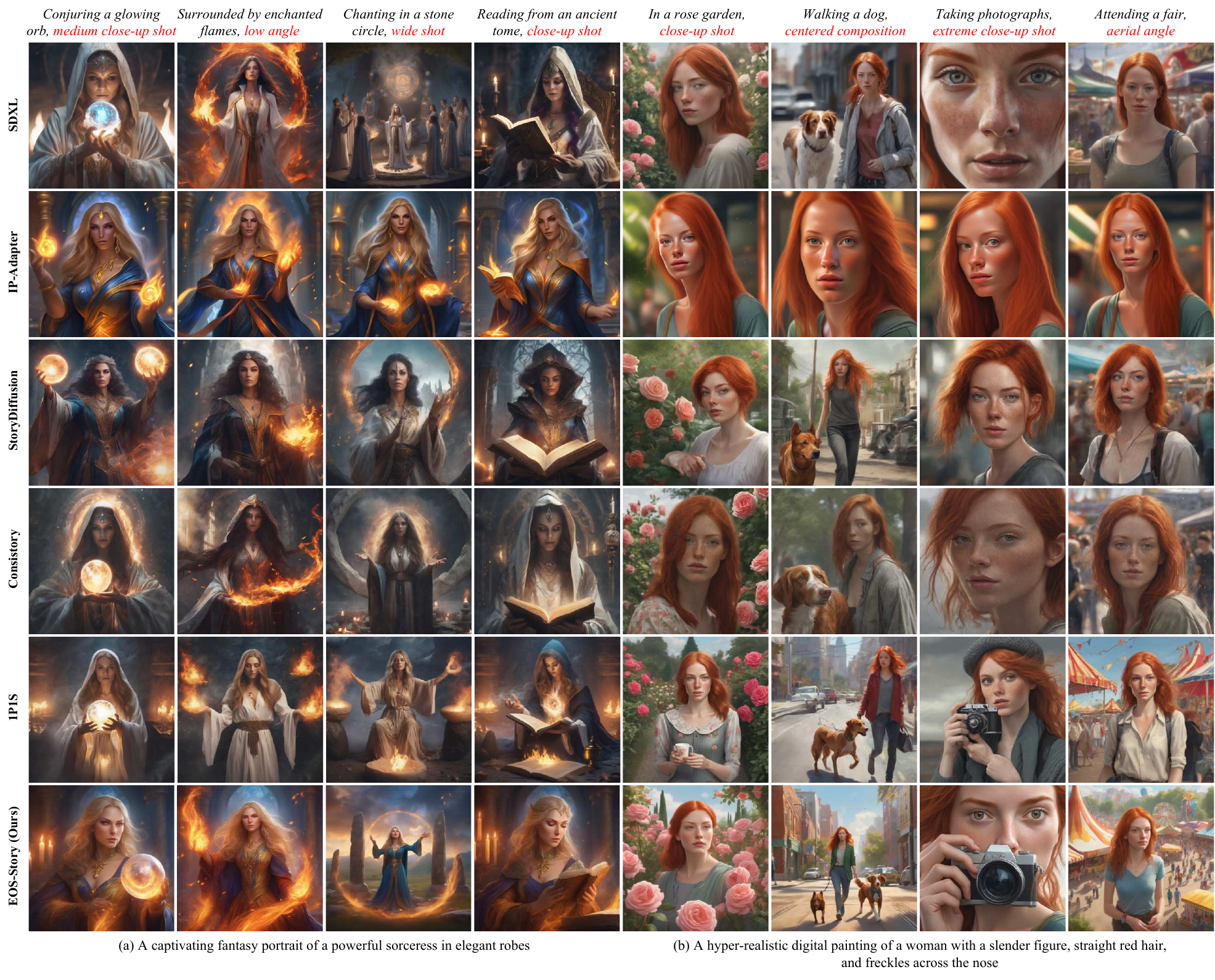}
    \caption{\textbf{Additional qualitative comparison on framing phrase adaptation.}}
    \label{fig:add_compare3}
\end{figure*}

\begin{figure}[!t]
\centering

\begin{minipage}[t]{0.48\textwidth}
    \centering
    \includegraphics[width=\linewidth]{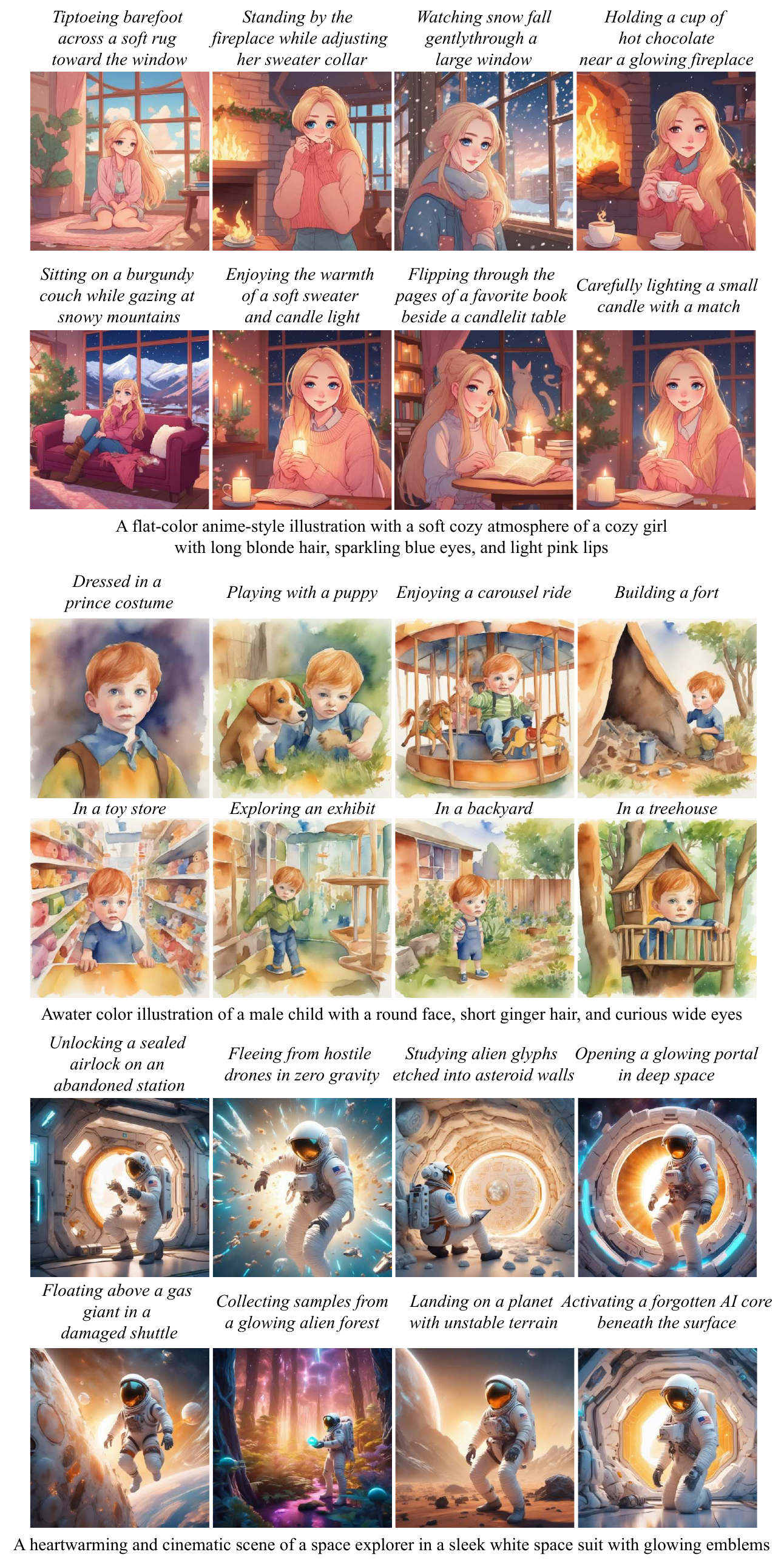}
    \captionof{figure}{\textbf{Medium length story generation.}}
    \label{fig:add_qual_midstory}
\end{minipage}
\hfill
\begin{minipage}[t]{0.48\textwidth}
    \centering
    \includegraphics[width=\linewidth]{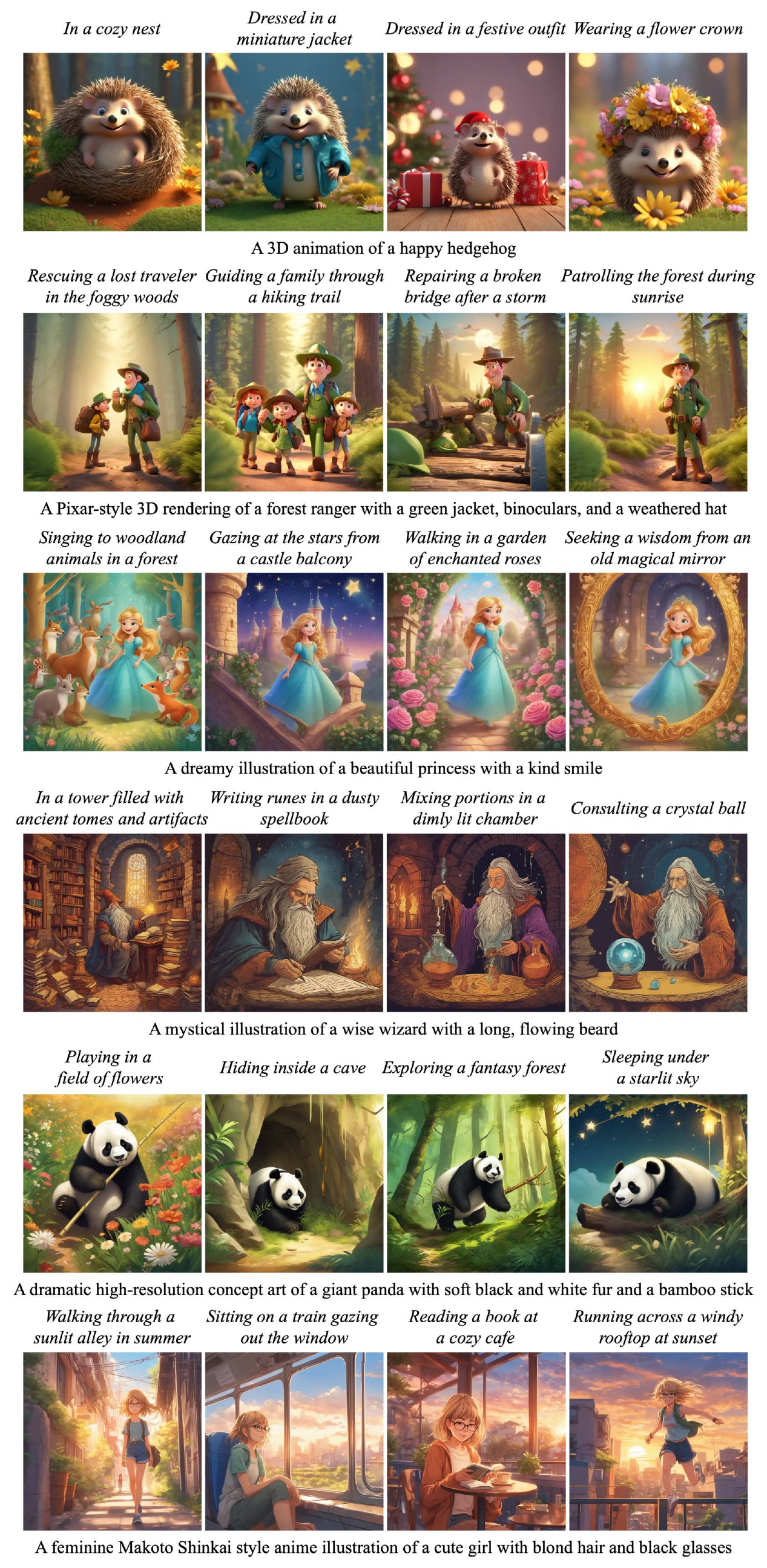}
    \captionof{figure}{\textbf{Short length story generation.}}
    \label{fig:add_qual_shortstory}
\end{minipage}

\end{figure}

\begin{figure*}[!t]
    \centering
    \includegraphics[width=1.0\textwidth]{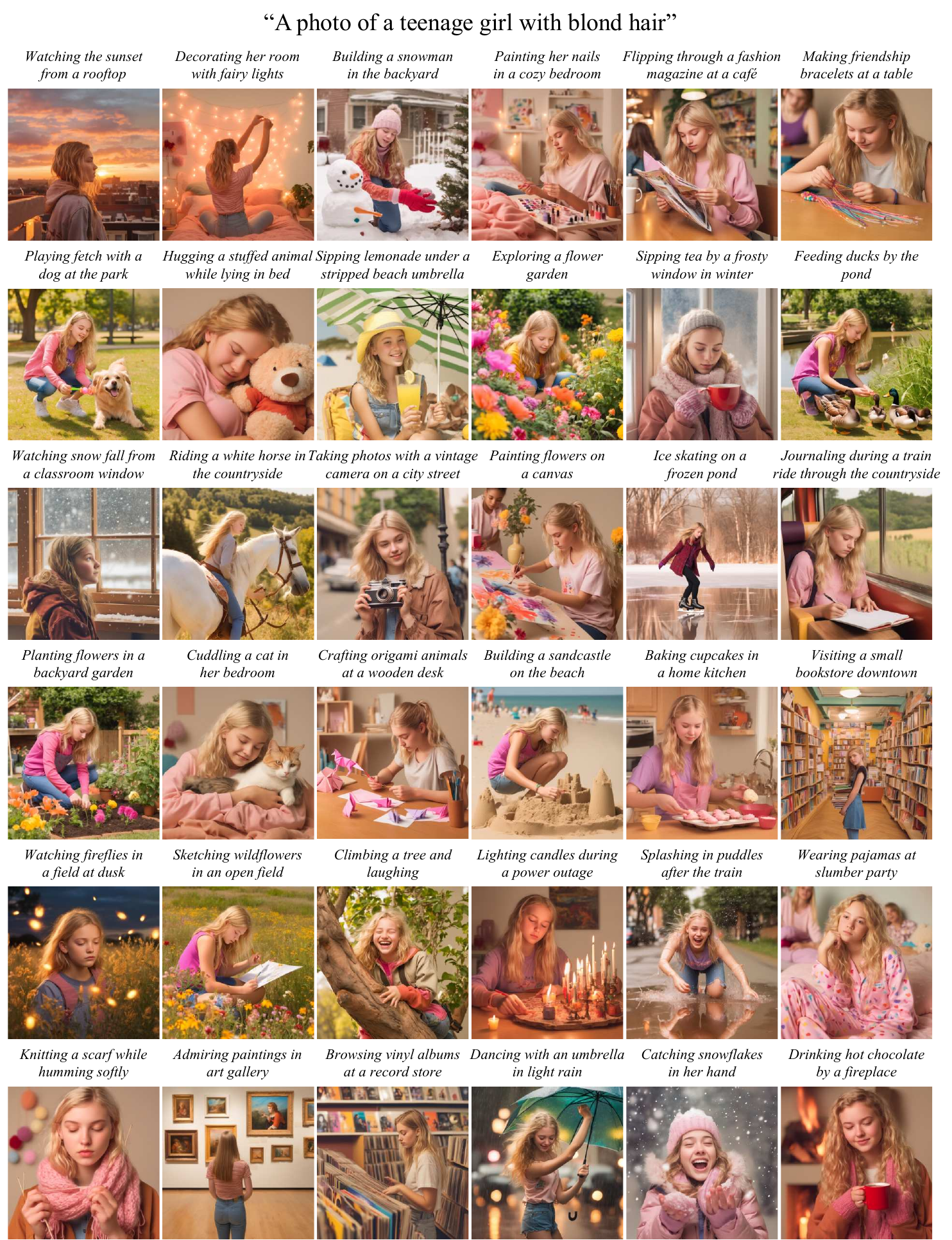}
    \caption{\textbf{Long story generation.} Our model accurately reflects detailed scene descriptions while preserving subject identity, even in extended scenarios.}
    \label{fig:add_qual_longstory}
\end{figure*}

\begin{figure*}[!t]
    \centering
    \includegraphics[width=1.0\textwidth]{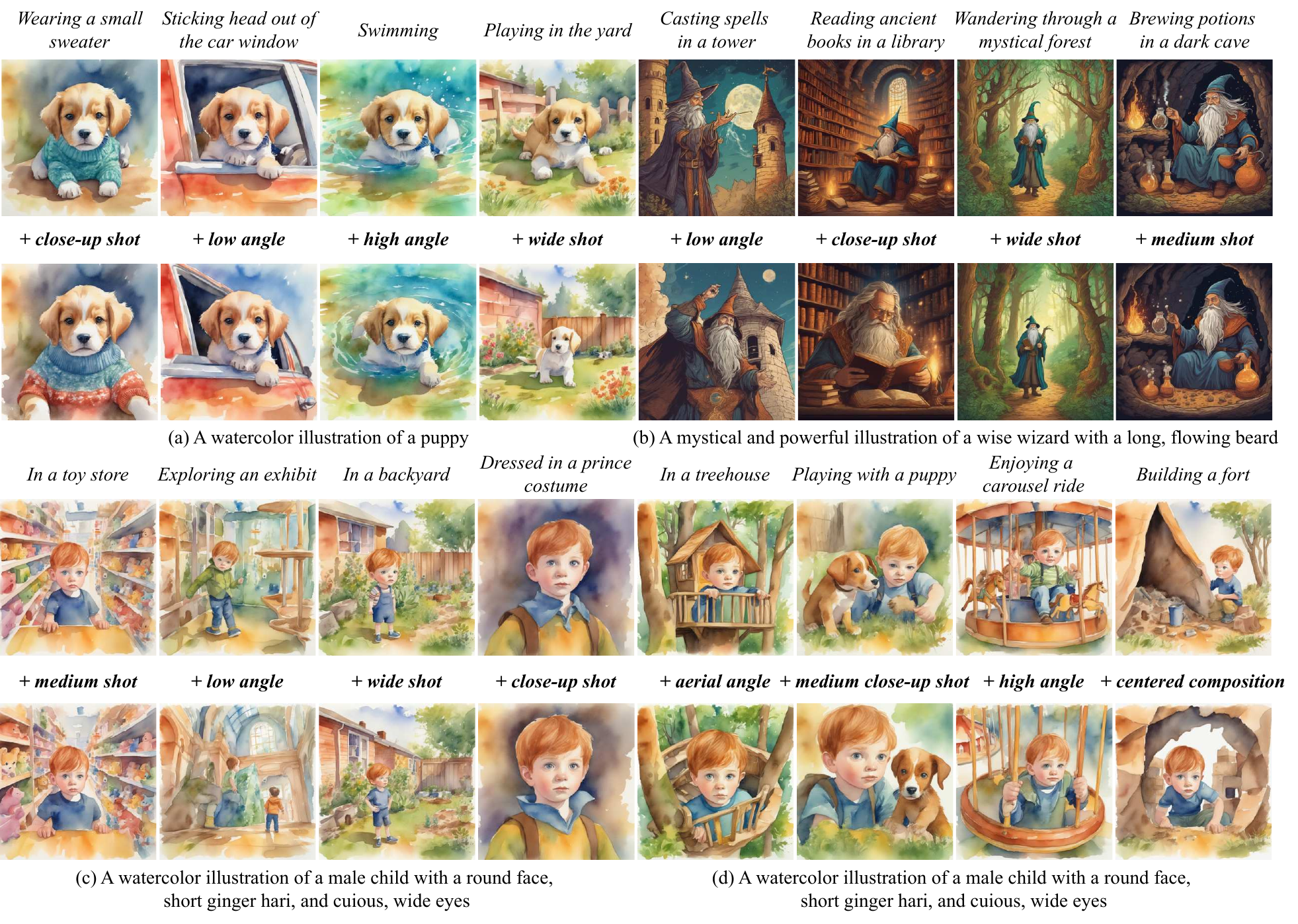}
    \caption{\textbf{Additional framing phrase adaptation results.}}
    \label{fig:add_qual_view}
\end{figure*}

We present additional qualitative comparisons in \cref{fig:add_compare2} and \cref{fig:add_compare3}.
As shown, EOS-Story effectively preserves subject identity while achieving better alignment with the given prompts—even in complex scenes involving diverse framing phrases.
Our method also supports generating storyboards of varying lengths (short, medium, long), as illustrated in \cref{fig:add_qual_midstory} and \cref{fig:add_qual_shortstory}, and \cref{fig:add_qual_longstory}.
Furthermore, \cref{fig:add_qual_view} demonstrates how our model adapts framing structures in response to different framing phrases.

\begin{figure}[!h]
    \centering
    \includegraphics[width=0.55\textwidth]{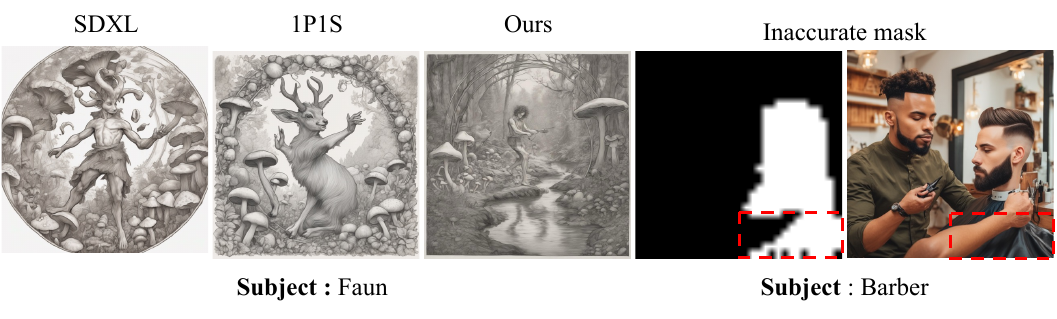}
    \caption{
    \textbf{Failure cases.} There are cases where the baseline model (SDXL) struggles to generate certain concepts (e.g., ``faun''). Additionally, since our method—like prior works—leverages the cross-attention map as a mask, the resulting masks sometimes capture irrelevant regions.}
    \label{fig:failure}
\end{figure}

\section{Limitation \& Future Work}
Although our model effectively balances prompt and subject coherence, several limitations remain.
While the cross-attention map generally captures subject regions accurately, it occasionally fails to attend to the correct area, leading to misaligned or partially misplaced generations—issues also observed in previous works.
This can be handled by explicitly using the subject mask as an additional condition.
In addition, there are cases where the model fails to generate accurate images at all (e.g., \cref{fig:failure} ``faun''), which appears to be an inherent limitation of the underlying SDXL backbone.
Furthermore, since Cross-Contextual Bridging exploits the contextual summarization property of the CLIP $\langle$EOS$\rangle$ token, it does not directly transfer to models based on T5 or LLM-based text encoders, where the $\langle$EOS$\rangle$ token no longer serves a comparable role. Enabling analogous residual-based semantic composition on such encoders is a promising direction for extending our framework to a broader range of modern diffusion models.


\end{document}